\documentclass[a4paper,fleqn]{cas-sc}
\usepackage[authoryear]{natbib}
\usepackage{graphicx}
\usepackage{blindtext}
\usepackage{float}
\usepackage{multicol}

\def\tsc#1{\csdef{#1}{\textsc{\lowercase{#1}}\xspace}}
\tsc{WGM}
\tsc{QE}
\tsc{EP}
\tsc{PMS}
\tsc{BEC}
\tsc{DE}
\begin{document}
\let\WriteBookmarks\relax
\def\floatpagepagefraction{1}
\def\textpagefraction{.001}
\shorttitle{Crime Prediction using Machine Learning}
\shortauthors{F. T. Shohan et~al.}
%\begin{frontmatter}

\title [mode = title]{Crime Prediction using Machine Learning with a Novel Crime Dataset}
\author[1]{Faisal Tareque Shohan*}[orcid=0000-0002-3872-5758]
\ead{faisaltareque@hotmail.com}
\address[1]{Dept. of Computer Science and Engineering, Ahsanullah University of Science and Technology, Dhaka, Bangladesh}
% \cortext[1]{Corresponding author}
\cortext[1]{Equal Contribution}
% \ead{@gmail.com}
\author[1]{Abu Ubaida Akash*}
\ead{akash.ubaida@gmail.com}

%\address[1]{Dept. of Computer Science and Engineering, Ahsanullah University of Science and Technology, Dhaka, Bangladesh}
\author[2]{Muhammad Ibrahim} [orcid=0000-0003-3284-8535]
\address[2]{Dept. of Computer Science and Engineering, Univeristy of Dhaka, Dhaka, Bangladesh}
\ead{ibrahim313@du.ac.bd}
\author[1]{Mohammad Shafiul Alam}
%\address[1]{Dept. of Computer Science and Engineering, Ahsanullah University of Science and Technology, Dhaka, Bangladesh}
\ead{shafiul.cse@aust.edu}

\begin{abstract}
Crime is an unlawful act that carries legal repercussions. Bangladesh has a high crime rate due to poverty, population growth, and many other socio-economic issues. For law enforcement agencies, understanding crime patterns is essential for preventing future criminal activity. For this purpose, these agencies need structured crime database. This paper introduces a novel crime dataset that contains temporal, geographic, weather, and demographic data about 6574 crime incidents of Bangladesh. We manually gather crime news articles of a seven year time span from a daily newspaper archive. We extract basic features from these raw text. Using these basic features, we then consult standard service-providers of geo-location and weather data in order to garner these information related to the collected crime incidents. Furthermore, we collect demographic information from Bangladesh National Census data. All these information are combined that results in a standard machine learning dataset. Together, 36 features are engineered for the crime prediction task. Five supervised machine learning classification algorithms are then evaluated on this newly built dataset and satisfactory results are achieved. We also conduct exploratory analysis on various aspects the dataset. This dataset is expected to serve as the foundation for crime incidence prediction systems for Bangladesh and other countries. The findings of this study will help law enforcement agencies to forecast and contain crime as well as to ensure optimal resource allocation for crime patrol and prevention.

%Every crime incident is recorded with place, time, weather, and demographic information. Hand-selected information from crime reports published in The Daily Star \cite{thedailystar}, makes up the crime data. Demographic information is sourced from the 2011 Bangladesh census \cite{bdstatistics2011}. Through statistical analysis, this paper also investigates the relationship between weather, crime, and demographic information. To forecast crime, a number of machine learning classifiers such as Decision Tree, Random Forest, XGBoost, Ada Boost, and Extra Tree classifiers has been applied. The Random Forest classifier outperformed others in forecasting crime. On a balanced dataset generated using SMOTE oversampling XGBoost and Extra Tree forecasts crime incidents with 59\% accuracy. This dataset could serve as the foundation for a crime incidence record system. The findings of this study can help police and other law enforcement agencies forecast and prevent future crime, as well as ensure optimal resource allocation for crime patrol and prevention.
\end{abstract}

\maketitle

\section{Introduction}
\label{sec:intro research questions}
Crime is a prevalent concern of any society. It has an impact on the quality of life and economic prosperity of a society. It is a critical factor in determining whether or not individuals should visit a city or country at a specific time or which areas they should avoid if they want to do so. It is also a key indicator of the development and social well-being of a country. Hence minimizing the crime activities has always been a priority of a government.

Law enforcement organizations leverage lawful means to contain the crime rate. Artificial intelligence, in particular, data science and machine learning disciplines have been offering tremendous benefits to almost every sector of a society including the law and order sector. The law enforcement agencies continue to seek advanced advanced information systems that employ state-of-the-art machine learning techniques to better safeguard their communities as crime rates are on the rise across the world. In many regions of the world, the domain of analyzing and detecting crimes using machine learning is gaining intensive research both from academia and industry.

Although crimes can occur anytime anywhere, criminals usually operate in their comfort zones and strive to recur the crime under similar circumstances once they are successful \cite{tayebi2012understanding}. This implies that crime incidents often leave trails of  consistent patterns. If this pattern is detected by the crime prevention authorities, then the crime may be preemptively stopped. Machine learning algorithms are being used to analyze and forecast crime, giving the security agencies a totally new viewpoint \cite{toppireddy2018crime}.

\subsection{Motivation and Research Questions}

As machine learning relies primarily on historical data to predict future outcomes, crime records from the past are necessary for analyzing and forecasting criminal incidents. These records must be in  structured and standard form in order to be used by machine learning algorithms. While in some countries this research is emerging (as detailed in Section~\ref{sec:related}), Bangladesh, despite having a sizable population of over 160 million, lacks any such study. To the best of our knowledge, no standard, structured, and day-to-day based crime data are available for Bangladesh. On Bangladesh Police's official website \cite{bdstatisticscrime}, some aggregated crime statistics are available to the general public. However, only the annual counts of a few crime types are provided there. Using these data, a Github repository containing a few types of crimes are available \cite{ibrahim_bdcrimestat}, but this dataset is neither systematically developed nor can it be considered extensive and standard. Also, the major factors that influence crime occurrence are not present in these data. In countries such as the United States \cite{sancrime}, \cite{chikagocrime}, the United Kingdom \cite{ukcrime}, and Canada \cite{canadacrime}, there are numerous widely used crime datasets available for researchers. Our research aims at minimizing this gap for Bangladesh. % However, there are no historical crime records containing spatio-temporal information in Bangladesh.
We attempt to address the following research question:

\vspace{0.2 cm}

``\emph{In order to facilitate prediction of crime occurrence in Bangladesh, how can we develop a standard machine learning dataset so as to apply supervised machine learning algorithms on it }?''

\vspace{0.2 cm}

The above question spawns a few sub-questions that include:
\begin{itemize}
    \item How to collect past crime data?
    \item How to extract information of crime incidents?
    \item How do climate and demographic information impact crime occurrence?
    \item How to select or engineer meaningful features for crime prediction?
    %\item How accurately can crime be predicted based on location, time, weather, and demographics?
    \item How to apply machine learning models to the crime data of Bangladesh?
    %Is the proposed dataset sufficient for crime prediction? Will a more robust dataset enable more accurate crime prediction?
\end{itemize}
%For this study, historical crime incident data is required in order to comprehend past crimes and extract underlying patterns from these crime events. Based on the discovered patterns, it is possible to predict future crimes.

\subsection{Contribution}

Below we list the major contributions made by this study:

\begin{itemize}
    \item We introduce a crime dataset with spatio-temporal, weather, and demographic information covering six types of crime incidents of Bangladesh over a seven year time span. Several factors that influence the crime occurrence are systematically extracted from a daily newspaper, a weather database, and Bangladesh National Census report. %A popular daily newspaper of Bangladesh is used as the source for the crime news, while the Bangladesh National, 2011 data is consulted for collecting demographic data. Weather data are collected using standard weather data providers.
    Thus, in our dataset, the impact of place, time, weather, and demographic data on crime occurrence are duly reflected.
    \item We apply several feature engineering methods to further improve the quality of the data and to make it suitable for machine learning tasks.
    \item We perform exploratory data analysis of the newly developed dataset and elicit useful patterns.
    \item Some popular supervised machine learning algorithms, namely Random Forest, XGBoost, AdaBoost, Extra Tree, and Decision Tree are employed to forecast criminal incidents based on our dataset. SMOTE oversampling technique is  also utilized to minimize data imbalance problem. We achieve reasonable accuracy in the prediction task. The findings of this study is expected to assist the law enforcement agencies in prediction, prevention of crimes in Bangladesh and to help in better resource allocation for crime patrol. %a particular location, time period, and demographic.
%The random forest classifier outperformed other methods for crime prediction. XGBoost and ExtraTree performed better than other classifiers on this balanced dataset.

\end{itemize}

The remaining portion of the paper is organized as follows:  Section~\ref{sec:methodology} describes the methodology of this study. Proposed dataset and its curating process is presented in detail in Section~\ref{sec:data aqcuisition basic features}. Section~\ref{sec:data processing feature enhance} describes the data processing steps and feature engineering methods. Section~\ref{sec:data analysis} performs exploratory data analysis on the newly developed dataset. Section~\ref{sec:ML models and results} discusses the experimental results after applying machine learning  models. Section~\ref{sec:related} relates this work to the existing research on crime forecasting. Section~\ref{sec:conclusion} brings the paper to a conclusion.

\section{Methodology}
\label{sec:methodology}
%There is no universal method for conducting research. Every study has its own workflow.

\begin{figure}[htbp]
% \begin{figure}
   \centering
   \begin{tabular}{@{}c@{\hspace{.5cm}}c@{}}
       \includegraphics[page=1,width=.75\textwidth]{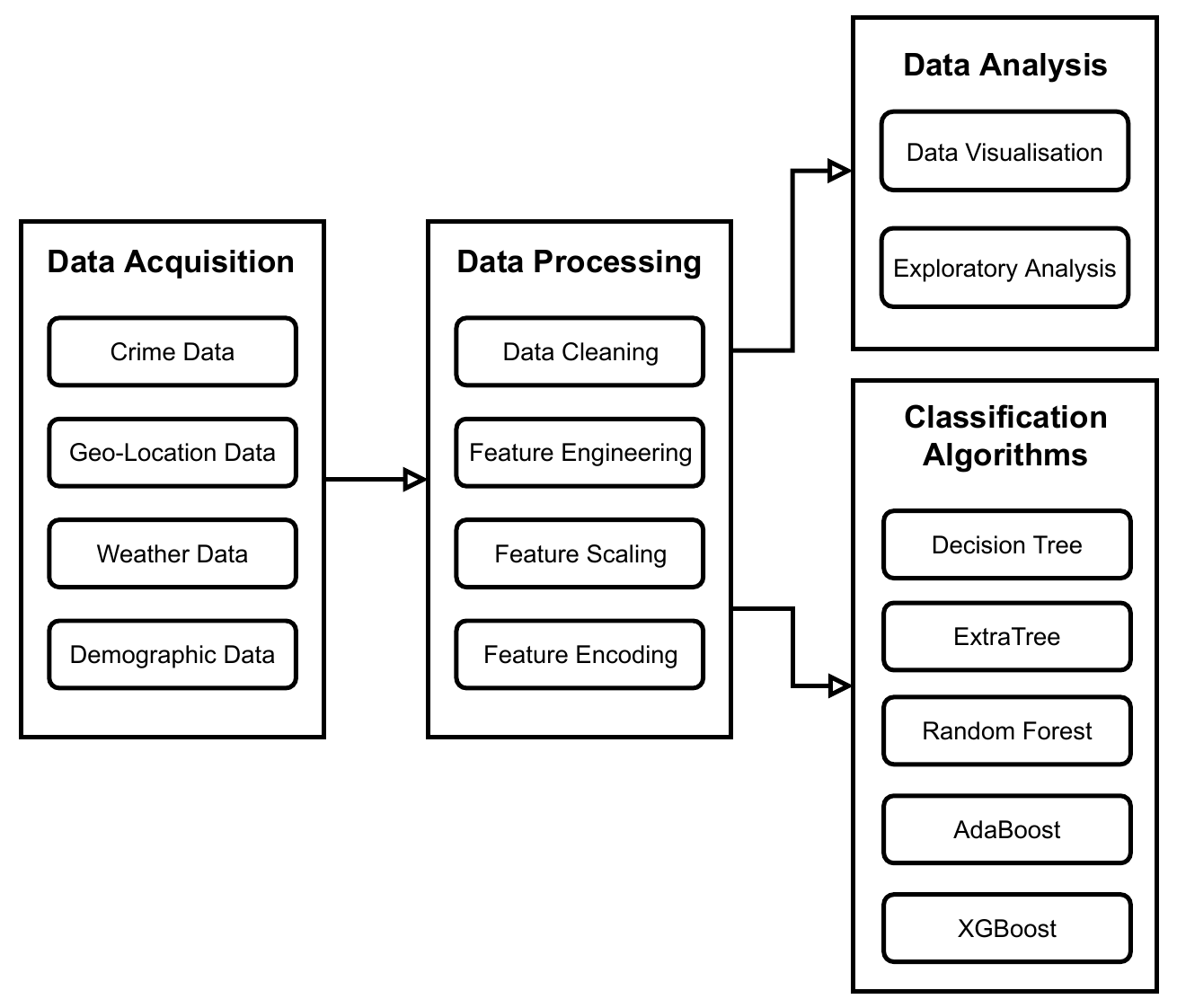}
   \end{tabular}
 \caption{Methodology of this research.}
 \label{fig:methodology}
\end{figure}

The workflow of this investigation can broadly be divided into four phases as depicted in Figure~\ref{fig:methodology}. Briefly, the phases are as follows:
\begin{enumerate}
    \item The first phase is data acquisition. We collect, both manually and automatically, crime news articles of the years 2013 -- 2019 from a popular daily newspaper of Bangladesh named ``The Daily Star'' \cite{thedailystar}. We also garner geographic, weather, and demographic information related to the collected crime incidents from different sources since these factors affect the crime occurrences. %At the end of this phase, we have 31 features in the dataset with 6574 crime instances. %Regardless of the field of research, data collection is typically the first and most crucial step in the research process. Depending on the type of information required, the methodology for collecting data differs across disciplines. This stage involves the gathering of demographic, weather, and crime data.
    \item The second phase is data processing and feature extraction. %Data processing is a crucial step in machine learning, as it improves the quality of the data to facilitate the extraction of insightful information from the raw data.
    Since the data contains raw text, cleaning this data is necessary. After cleaning the data using standard tools, we derive more crime features from the text. We also apply feature engineering techniques to further improve the quality of the dataset, thereby make the dataset ready for being fed into machine learning prediction algorithms. This way we develop the dataset which we call \emph{CrimeDataBD} -- the  first-ever standard crime dataset of Bangladesh containing 6574 crime instances and 36 features.
    \item The third phase involves analysis of the newly developed dataset using an exploratory approach. The goal here is to better understand the characteristics of the dataset before applying the machine learning models.
    \item Finally, in the fourth phase we apply several machine learning classification algorithms on the dataset and analyze the results. %For classification, there are many machine learning algorithms. %Both supervised and unsupervised strategies can be used for classification task.
    We employ five supervised algorithms, namely, Decision Tree, Extra Tree, % has been applied on the data. Additionally,  ensemble learning classifier like
    Random Forest, Adaboost, and XGBoost.  The prediction accuracy produced by the classification algorithms are then analyzed and found to be satisfactory. %Figure~\ref{fig:methodology} demonstrates the entire research methodology of this study.

\end{enumerate}

The following sections describe these four phases in more details.

\section{Data Acquisition}
\label{sec:data aqcuisition basic features}

%Data collection is the process of gathering, measuring, and analyzing accurate information for research using standard and validated methods.
This section describes how we collect the crime data and prepare it for the next stages. %the BD-Crime-Demographic dataset and its curation process in details.
The overall process of data acquisition is administered as follows: First, we select a good number of crime news articles from the newspaper (this process will be elaborated shortly) and categorize the articles into six type of crimes. Second, for each of the crime incidents of all six categories, we extract various information about the crime such as date, place, time, victim's information, criminal's information etc. This comprises the basic dataset. Third, among the information extracted in the previous step, we choose the features (such as crime place, time etc.) that are able to assist in predicting future occurrence of crime incidents. Here we also derive some other features which are not present in the news articles such as weather information of the crime occurrence time and demographic information of the area of the crime. Thus, we get a dataset that contains various useful information for predicting future crime occurrences.

Regarding the first step mentioned above, the following steps are taken in order to fetch the crime incidents from the daily newspaper ``The Daily Star'' and to extract features from the articles:

\begin{enumerate}
  \item We read the Front, Back, City, and Country sections of the newspaper archive covering the years 2018 and 2019. After skimming through around twenty eight thousand news articles, two thousand potential crime news articles are shortlisted and divided into six crime categories which are murder, rape, assault, robbery, kidnapping, and body-found.
  \item We then extract some keywords for each category of crime from  the 2000 articles. We do this by first tokenizing (i.e., breaking down into words aka tokens) the headlines of these articles. We then select the tokens having the most frequency of occurrence in a category. We exclude some irrelevant keywords using manual judgement.
  \item Our next step is to automate the process of fetching even more crime news articles. We do this by using a Web Crawler tool\footnote{A web crawler is a tool that automatically fetch URL links from Internet. We choose \emph{BeautifulSoup4} \cite{bs4} as the web crawler.}. We crawl, without mentioning any criteria, around sixty thousand news links and headlines from the aforesaid four sections of the newspaper archive for the duration of the years 2013 -- 2017. Definitely all articles of this huge collection are not crime news, so we then use the keywords collection (developed in the previous step) to fetch the crime news among these 60,000 articles. Here we employ another tool called FuzzyWuzzy \cite{fuzzywuzzy}. This way we retain around 4,700 crime news articles among the 60,000 ones. Added to these are the previously collected 2000 articles. Some duplicate articles are manually discarded. Thus the final number of crime news articles stands at 6574. At this stage, the collection is ready for feature selection and extraction.
\end{enumerate}

In the remainder of this Section we further elaborate the above-mentioned steps.

\subsection{Crime News Source Selection}
Since every newspaper covers crime news, it is a great source for obtaining information about past criminal activities. %A newspaper is sought in which previous crime-related news can be found.
The Daily Star \cite{thedailystar} is selected as the source for crime news since it is Bangladesh's one of the most widely read daily English-language newspapers. To avoid news duplication, we work with only one newspaper as different newspapers cover the same crime incident.

\subsection{Basic Data Acquisition}
%The Daily Star publishes numerous articles about crime.
Among the published crime news, the most prevalent types of crime news are found to be murder, rape, assault, robbery, kidnapping, and corruption. The frequency of corruption-related crimes is lower than that of other types of crime. In contrast, the number of reports containing discovery of an unknown person's body is found to be notably high. Even though body-found news is classified as murder news, we notice that this category of report usually lacks a significant amount of the information that are found in murder news. Therefore, we classify body-found as a separate type of crime.

%\subsubsection{Criteria for Selecting News}
%Certain criteria are followed throughout the collection of crime incident news.
The selected news may include articles about murder, rape, assault, robbery, kidnapping, or body-found. In addition, if an incident involving multiple crimes is reported in the news, each sort of crime is identified. War crimes, accidents, deaths or injuries resulting from landslides, road collisions, gunfights, police shootings, human trafficking etc. are disregarded.

%\subsubsection{News Search}
%The Daily Star newspaper website publishes news in different sections.
News about crime incidents are usually found on Front, Back, Country, and City sections in the Archive portal. At the time of data collection, the Archive portal \cite{dailystararchive} of The Daily Star was publicly accessible. Unfortunately, the Archive  portal is not accessible anymore. However, all the collected news are still accessible through the direct links (which is provided in the dataset, like this one: \cite{examplenews}).

\subsubsection{News Link Collection using Manual Process}
Since The Daily Star does not offer a news API, at first we resort to a manual process for acquiring data. We begin our news link collection with the news published during 1st January, 2019 -- 31st December, 2019 found in the Archive portal. After reading the headline and skimming through the content of the report, we determine if the news is about crime and to what category it belongs. Every news article on the Front Page, Back Page, City, and Country is reviewed to identify crime-related news. After reviewing around 28000 archived news articles, 2000 crime-related news links are selected.

\begin{figure}[htbp]
    \centering
    \includegraphics[width=100mm,scale=0.9]{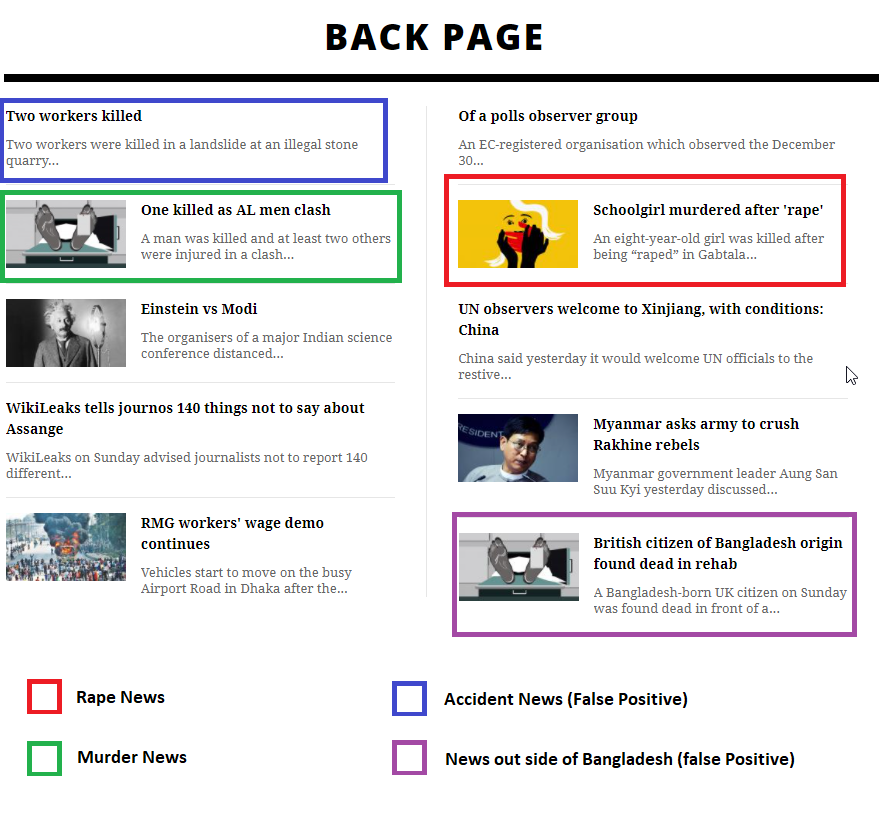}
    \caption{An example crime and non-crime news from the Backpage of the Daily Star.}
    \label{fig:crimenewsbackpage}
\end{figure}

Sometimes it is possible to misinterpret a non-criminal news as a criminal one if only headline is read. For instance, the headline ``Two Workers Killed'' suggests a crime has happened. However, a thorough reading of the news reveals that it, in fact, reports an accident. Such an instance is illustrated in Figure~\ref{fig:crimenewsbackpage}. This shows that collecting crime-related news by manually reading every news report from the four newspaper sections was quite cumbersome and time-consuming process.

\subsubsection{News Link Collection using Web Crawler}
Since The Daily Star lacks a news API, we use Web Crawling as the next stage. To automatically gather more news links, we construct a web crawler using a tool called \emph{BeautifulSoup} \cite{bs4}. Using this tool we traverse all articles published from January 1, 2013 to December 31, 2017 from the Front Page, Back Page, City, and Country sections from the archive portal of The Daily Star and gather all headlines and links. This way approximately 60,000 news links are collected.  %It is laborious and time-consuming to choose crime news after reading every headline and article's substance. To remedy this,
%These articles are further iltered for further processing.

\subsubsection{Filtering Crawled News Collection to Identify Crime News}
Obviously not all of the 60,000 crawled articles report crime news. Now we need a method to automatically select crime news from this vast collection. While manually gathering news links, we analyze a headline to see whether there are any significant cues within the headlines about crime news. For instance, the words `harass', `attacked', `assault', `torturing', `stabbed', `beat', `brutally', `shoot acid', `molested', `burn', `shot', `stabbing', `chained', `tied', `injury', `harassed', `abuse', `brutalised', `forcibly', `cuts' etc. indicate the assault crime. Other keywords are discovered by analyzing headlines from the previously (manually) collected 2000 crime news. Here we automate the process of calculating frequencies of all words of all the 2000 headlines, and then manually select a group of keywords from each crime category. These keywords are given in Table~\ref{Tab:keywordineachcategory}. Using the FuzzyWazzy \cite{fuzzywuzzy} tool for string matching, we then utilize these keywords to filter the crime news from the 60000 news links. More intervention from us was necessary to identify and remove news about accidents, natural disasters, war crimes, human trafficking, and news from outside Bangladesh. This way 4700 news articles are selected as crime news from the 60000 ones.

% We use FuzzyWuzzy \cite{fuzzywuzzy} string matching library which matches the keywords of a category against a headline, and thus puts each headline into one of the six categories

\begin{table}[width=.95\linewidth,cols=2,pos=h]
\caption{Keywords in each Crime Category}
\begin{tabular*}{\tblwidth}{@{} LL@{} }
\toprule
Category & Keywords\\
\midrule
Murder & murder, murdered, kill, killed, homicide, slaying, manslaughter, shoot, dead, assassinate, \\ & stabbed,  suffocated, poisoned  \\
\\
Rape & raping, rape, raped, gang-raped, rapes, rapist, gang-rape \\
\\
Assault & harass, attacked, assault, attack, torturing, tortur, stabbed, beat,
brutally, unconscious,  \\ & forced, boiling, shoot, acid, thrown, molested,
assaulted, burn, assaults, shot, stabbing, tied, \\ & chained,  brutality, sexually,
injury, harassed, abuse, brutalised, assailant, brutalise, attacks, \\ & forcibly, cuts,  stalking, sexual,  molestation, shave, throwing, cruelty, caned, wrath, abusing,  \\ & burnt, hack, molest, mercilessly, resists, stab \\
\\
Robbery & mugger, mugged, robber, loot, snatch, robbed, robbery, looted  \\
\\
Kidnap & abduction, abduct, abducted, abducting, kidnap, rescued, missing, traceless \\
\\
Body Found  & found body, body found, bodies found, body recovered, bodies recovered,  found dead, \\ & found murdered, found hanging,  found bodies, murdered found,  dead found, recovered body, \\ & recovered bodies,  hanging body, hanging bodies,  bullet-hit body , bullet-hit bodies, \\ & decomposed body, decomposed bodies  \\
\bottomrule
\end{tabular*}
\label{Tab:keywordineachcategory}
\end{table}

\begin{table}[width=.7\linewidth,cols=2,pos=h]
\caption{Distribution of crime instances in each crime category of the dataset.}
\begin{tabular*}{\tblwidth}{@{} LLLL@{} }
\toprule
Category & Data\\
\midrule
Murder & 1518 \\
Rape & 1193 \\
Assault & 1097 \\
Robbery & 598 \\
Kidnap & 651 \\
Body Found & 1517 \\
\bottomrule
\end{tabular*}
\label{Tab:Dataineachcategory}
\end{table}

%\subsubsection{Obtained Crime Data}
%Information extraction from gathered news articles yields the collection of 6700 data points.
Note that a few of these news were found to be duplicate news, we manually identified and eventually discarded them. Finally, 6574 crime incidents are recorded in our dataset that occurred during the time span of 2013 -- 2019. The data collected for each category of crime are listed in Table \ref{Tab:Dataineachcategory}.

%\subsubsection{What information could be extracted from news?}

\subsection{Basic Feature Preparation for Machine Learning}
Now that we have gathered the crime news reports in natural language format, it is time to extract useful information from these so that a machine learning model can utilize this dataset to predict future crime occurrences. Machine learning algorithms need some features that can signal about the possibility of future crimes. In this subsection we elaborate this feature preparation process.

%After collecting the news articles, we move our attention to machine learning setup where we, in order to predict future crime incidence.

\subsubsection{Information Extraction from News Reports}
We carefully inspect every category of crime news to find out what information about the crime is described in the news. Most common information found in a crime news article are:

\begin{multicols}{2}
\begin{enumerate}[i)]
\item  News date
\item Crime approach
\item Relation between victim and criminal
\item Incident place
\item Incident time
\item Victim profession
\item Victim age
\item Victim's address
\item Criminal's age
\item Motive behind the crime
\item If criminal is arrested or not
\end{enumerate}
\end{multicols}

In addition, news articles may include information such as the victim's or offender's political and religious involvement. However, not all crime-related news articles contain all these fields. Therefore, we manually extract these information from all the news articles. For each type of crime, the following information are extracted:

\textbf{Murder:} news date, incident date, if arrested, part of the day of incident, incident place, murder approach, murder weapon, motive,  victim age, victim gender, victim profession, victim religion, victim address, criminal age, criminal gender, criminal profession, criminal religion, relation between victim and criminal.

\textbf{Rape:} news date, incident date, if arrested, part of the day of incident, incident place, no of victims, victim age, victim gender, victim profession, victim religion, victim address, criminal age, criminal gender, criminal profession, criminal religion, relation between victim and criminal.

\textbf{Assault:} news date, incident date, if arrested, part of the day of incident, incident place, motive,  victim age, victim gender, victim profession, victim religion, victim address, criminal age, criminal gender, criminal profession, criminal religion, no of criminal, relation between victim and criminal.

\textbf{Robbery:} news date, incident date, if arrested, part of the day of incident, robbed area, incident place,  victim state, victims injured, criminal age, criminal gender, criminal religion, no of criminal.

\textbf{Kidnap:} news date, abduction date, rescue date, if arrested, part of the day of incident, incident place, rescued place,  victim age, victim gender, victim profession, victim religion, victim address,, victim injured, criminal age, criminal gender, criminal profession, criminal religion,no of criminal, relation between victim and criminal.

\textbf{Body Found:} news date, incident date, part of the day of incident, incident place, victim age, victim gender, victim profession, victim religion, victim address, body state.

%It may be noted here that information extraction from the news was challenging. Reading each article and entering the data into the corresponding feature was time-consuming and laborious. %%However, there is no alternative possibilities that might have yielded the required outcome. Scraping this information would be virtually difficult. Even if scraping is used to collect information from news articles, each one had to be verified individually.

\begin{figure}[htbp]
    \centering
    \includegraphics[width=80mm,scale=0.7]{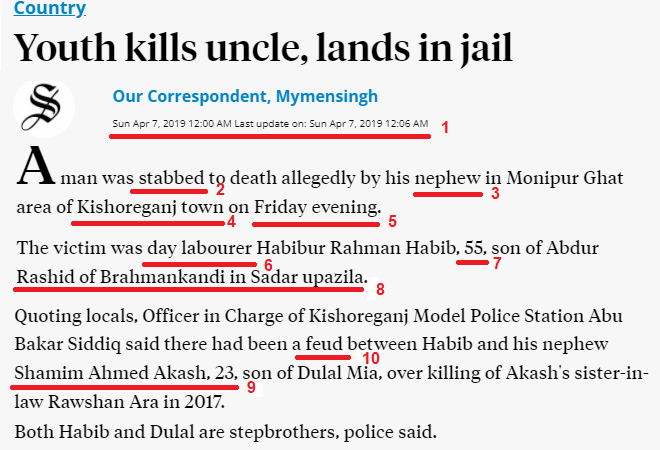}
    \caption{Example of Murder News Part 1}
    \label{fig:murdernewsexample1}
\end{figure}

\begin{figure}[htbp]
    \centering
    \includegraphics[width=80mm,scale=0.7]{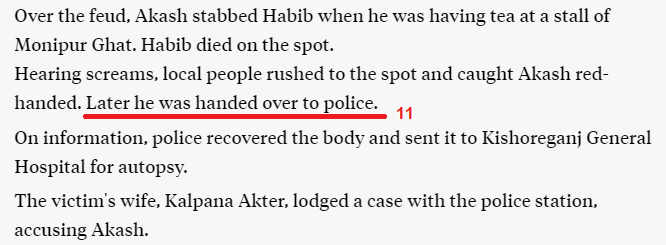}
    \caption{Example of Murder News Part 2}
    \label{fig:fig:murdernewsexample2}
\end{figure}

%We manually extract all these information from all the crime news articles.
We note that this manual information extraction process was quite cumbersome and time-consuming. In Figures~\ref{fig:murdernewsexample1} and \ref{fig:fig:murdernewsexample2} we show an example of this information extraction process.

\subsubsection{Feature Identification for Machine Learning}
While the dataset built so far is informative, it is not yet ready for machine learning tasks. One of the goals of the study is to forecast criminal activity. Before predicting a crime incident, it is necessary to identify the underlying pattern. %The previous section provides a detailed explanation of the method used to collect crime data. There are some specific details for certain crime category. However,

If we carefully analyze the information fields mentioned in the previous discussion, we see that incident date, part of the day of incident, and incident location may hint about the pattern of the crime. Fortunately, these three pieces of information are present in all crime news. Algorithms may be able to predict crime incidents with the help of the useful patterns hidden in these spatio-temporal data. So we select these three information as features for machine learning. Table~\ref{Tab:rawfeaturenews} lists these features.  %(Detailed description of these features are given later in Section~\ref{sec:data processing feature enhance}.)

\begin{table}[width=.7\linewidth,cols=2,pos=h]
\caption{Selected features for crime prediction from each crime incident.}
\begin{tabular*}{\tblwidth}{@{} LLLL@{} }
\toprule
Feature & Type\\
\midrule
Incident date           & Date                  \\
Incident place          & Categorical           \\
Part of the day of the incident        & Categorical           \\
\bottomrule
\end{tabular*}
\label{Tab:rawfeaturenews}
\end{table}

\subsection{Features Derived from Basic Ones}
While the above three features, i.e., incident date, incident place, and part of the day, may predict the chance of occurring a crime at a particular place in a particular timeframe, it is, from the machine learning perspective, better to have more information that may affect the crime occurrence. Therefore, using these basic features we derive three additional types of features, namely, geo-location features, weather features, and demographic features. Below we describe each of these.

\subsubsection{Geo-location Data Acquisition}
We obtain the geographic coordinates of a given location using the Map-box \cite{mapbox} Geocoding API from the Search service. Later, we add latitude and longitude for addresses.

For some addresses, Mapbox does not provide a response in terms of latitude and longitude. For example, even though the official name of a district is ``Chapainawabganj'', Geocoding services refer to it as ``Nawabganj''. Another example is, the district airport is still known as ``Jessore Airport'', even though the official name has been changed to ``Jashore'' \cite{districtnamechange}. This kind of inconsistency led to some oddities which we had to manually address.

\begin{figure}[htbp]
    \centering
    \includegraphics[width=100mm,scale=0.9]{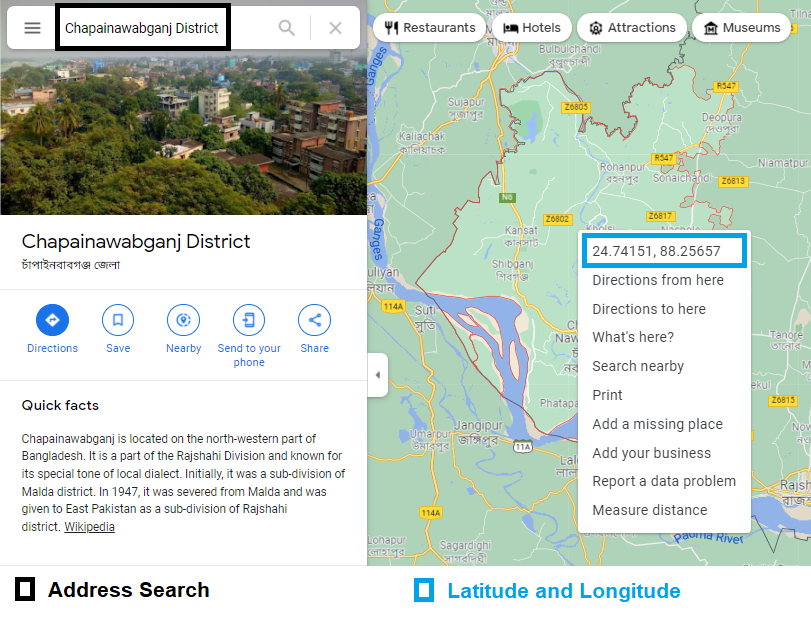}
    \caption{Collecting Latitude and Longitude from Google Map.}
    \label{fig:manualgeocoding}
\end{figure}

Google Map\footnote{\url{https://www.google.com.bd/maps}} is used to search for incident areas where no response or an incorrect response is received from Mapbox. A Google Map search gives us the latitude and longitude. Figure~\ref{fig:manualgeocoding} illustrates an example of this process. Table~\ref{Tab:weatherfeatures} lists the two weather features.

\begin{table}[width=.7\linewidth,cols=3,pos=h]
\caption{Geo-Location Features Derived from Incident Place (cf. Table~\ref{Tab:rawfeaturenews}).}
\begin{tabular*}{\tblwidth}{@{} LLLL@{} }
\toprule
Feature & Type & Derived From \\
\midrule
Latitude              & Numerical         & Incident Place    \\
Longitude             & Numerical         & Incident Place    \\
\bottomrule
\end{tabular*}
\label{Tab:geo-derivedfeaturenews}
\end{table}

\subsubsection{Weather Data Acquisition}
It is reasonable to assume that weather and criminal activity have some sort of association. Moreover, Heilmann and Kahn \cite{NBERw25961} show that on days with maximum daily temperatures being above 85 degrees Fahrenheit (29.4 degrees Celsius) the overall crime rises by 2.2 percent, and violent crime rises by 5.7 percent. Among various weather information, we select the ones that have a direct affect on crime such as visibility, cloudiness, temperature etc.

The Weatherstack \cite{weatherstack} API delivers accurate weather data -- both real-time and historical -- for a particular time and place.\footnote{However, access to the historical weather API is not included in the free version of Weatherstack, and so we purchased the paid version.} We collect the following information: temperature, weather type (encoded as a numerical code), precipitation, humidity, visibility, cloud coverage, heat, and season. %The necessary information from by a successful API call are Maximum Temperature, Minimum Temperature, Average Temperature, Weather Code, Weather Description, Precipitation, Humidity, Visibility, Cloud Coverage and Heat Index.
Table~\ref{Tab:weatherfeatures} shows the full list of the weather features.

\begin{table}[width=.7\linewidth,cols=3,pos=h]
\caption{Weather features collected from WeatherStack API (cf. Tables~\ref{Tab:rawfeaturenews} and \ref{Tab:geo-derivedfeaturenews}).}
\begin{tabular*}{\tblwidth}{@{} LLLL@{} }
\toprule
Feature & Type  & Derived From \\
\midrule
Max Temp              & Numerical      & Latitude, Longitude, Incident Date   \\
Avg Temp              & Numerical      & Latitude, Longitude, Incident Date    \\
Min Temp              & Numerical      & Latitude, Longitude, Incident Date   \\
Weather Code          & Categorical    & Latitude, Longitude, Incident Date   \\
Precipitation         & Numerical      & Latitude, Longitude, Incident Date   \\
Humidity              & Numerical      & Latitude, Longitude, Incident Date   \\
Visibility            & Numerical      & Latitude, Longitude, Incident Date   \\
Cloudcover            & Numerical      & Latitude, Longitude, Incident Date   \\
Heatindex             & Numerical      & Latitude, Longitude, Incident Date   \\
Season                & Categorical    & Latitude, Longitude, Incident Date   \\
\bottomrule
\end{tabular*}
\label{Tab:weatherfeatures}
\end{table}

\subsubsection{Demographic Data Acquisition}
Demographic data refers to information that is socioeconomic in nature and represents the characteristics of a geographic location. These data include population, household size, education, literacy rate etc. Demographic factors are crucial in understanding the crime rates of a place. The crime rate of an area are influenced by them. For example, the higher the literacy rate, the lower the possibility of crime; the higher the number of playgrounds, the the lower the possibility of crime; the lower the number of police stations, the higher the possibility of crime etc.

The National Census of Bangladesh was conducted in 2011 by the Bangladesh Bureau of Statistics \cite{bdstatistics} and the reports are made publicly available \cite{bdstatistics2011}. We manually extract these census data on population size, households, sex and age distribution, marital status, economically active population, literacy and educational attainment, religion, etc. for all of Bangladesh's districts and Upazilas\footnote{In Bangladesh, several Upazilas comprise a district, and several Unions comprise an Upazila.}, as well as for the major cities. Table \ref{Tab:demofeatures} shows the full list of features extracted from Bangladesh National Census report.

At this stage, we have 31 features in our dataset which are mentioned in Tables~\ref{Tab:rawfeaturenews}, \ref{Tab:geo-derivedfeaturenews}, \ref{Tab:weatherfeatures}, and \ref{Tab:demofeatures}.

\begin{table}[width=.7\linewidth,cols=3,pos=h]
\caption{Demographic features collected from Bangladesh Census Data, 2011 (cf. Table~\ref{Tab:rawfeaturenews}).}
\begin{tabular*}{\tblwidth}{@{} LLLL@{} }
\toprule
Feature & Type & Derived From \\
\midrule
Household Number      & Numerical     & Incident Place       \\
Male Population       & Numerical     & Incident Place       \\
Female Population     & Numerical     & Incident Place         \\
Total Population      & Numerical     & Incident Place       \\
Gender Ratio          & Numerical     & Incident Place       \\
Avg House Size        & Numerical     & Incident Place        \\
Population Density    & Numerical     & Incident Place         \\
Literacy Rate         & Numerical     & Incident Place         \\
Religious Institution & Numerical     & Incident Place         \\
Playground            & Numerical     & Incident Place        \\
Park                  & Numerical     & Incident Place       \\
Police Station        & Numerical     & Incident Place          \\
Cyber Cafe            & Numerical     & Incident Place       \\
School                & Numerical     & Incident Place         \\
College               & Numerical     & Incident Place          \\
Cinema                & Numerical     & Incident Place         \\
\bottomrule
\end{tabular*}
\label{Tab:demofeatures}
\end{table}

\section{Data Processing and Feature Enhancement}
\label{sec:data processing feature enhance}

In machine learning community, it is believed that the real predictive power of these astonishing mathematical techniques oftentimes lies in the quality and quantity of the data they work on. As we build our dataset from real-world text data, we need to apply a variety of methods to clean the dataset from noise and to standardize the dataset so it becomes ready-to-use for machine learning.

\subsection{Processing and Transforming Crime Data}
%Real-world data is frequently incomplete, inconsistent, lacking in certain behaviors or trends, and prone to a variety of errors. Preprocessing data is a tried-and-true solution to such issues.
% Raw data is preprocessed to make it ready for further processing.
Real world data is usually inconsistent, incomplete, and prone to a variety of errors. This situation aggravates when the data is text of natural language. In our case, since we develop the dataset from scratch, i.e., from a daily newspaper, emphasize on data cleaning and feature engineering is imperative to make it eligible for machine learning model fitting. In this sub-section we describe how we process and cure the three basic features, namely, incident place, incident date, and part of the day of the incident.

\subsubsection{Incident Place}
Incident locations are presented in news articles in a variety of formats. %When collecting data, these are added to the dataset in the order of their discovery.
In some cases, only the district of crime is mentioned, while in some other cases, both Upazila and the district information are available, and in some other cases, only the Union information is available. In our dataset, for each crime incident we keep all three fields but leave blank if any field is missing. %To resolve this inconsistency, we devise the following strategy:
%\begin{itemize}
%\item  Each incident place is split into three sections. A comma separates the Union or specific place name, the Upazilla name, and the District name.
%\item If a specific location or Union Name is not found, that location only has Upazilla and District fields. If Upazilla is not found, that location only has Union and District. If some cases in only contains District as Union or Upazilla are not mentioned in the news.
%\end{itemize}

The spellings of the Union, Upazila, and District names are collected from Wikipedia \cite{upazila} and Bangladesh government websites \cite{portalgovbd}. Some of these spellings, however, are found to be inconsistent with the official names. Some examples of such inconsistencies are shown in Table~\ref{Tab:SpellingDifference}.

\begin{table}[width=.7\linewidth,cols=2,pos=h]
\caption{Examples of Spelling Inconistencies in District Names.}
\begin{tabular*}{\tblwidth}{@{} LLLL@{} }
\toprule
Spelling of District in News Article & Official Spelling of District\\
\midrule
Netrakona & Netrokona \\
Jessore & Jashore  \\
Bogra & Bogura    \\
Barishal & Barisal  \\
Cumilla & Comilla    \\
Jhenaidha  & Jhenaidah   \\
Laxmipur  & Lakshmipur  \\
Chittagong   & Chattogram   \\
B'baria  & Brahmanbaria   \\
Moulovibazar   & Moulvibazar   \\
\bottomrule
\end{tabular*}
\label{Tab:SpellingDifference}
\end{table}

Since manually correcting these misspellings of place names is inefficient and time-consuming, we opt for string matching and replacement tools. The FuzzyWuzzy \cite{fuzzywuzzy} string matching module is used to correct all misspellings present in the news articles. FuzzyWuzzy is a library of Python which we use for string matching. %It uses Levenshtein distance to calculate the differences between sequences.

The majority of the crime occurrences include Upazila and District information, but do not include Union or a specific location name. As a result, having only the Upazila and District names for the incident location seems to be a reasonable choice. However, some of the news articles do not mention the Upazila or Union names. Instead, they use other location information like village, bus station, university, town, school, police station, hospital, market etc. We manually search all of these place names in Google search engine and extract the corresponding Upazila or municipal area. We then substitute those names with corresponding Upazila or municipal area names. The process, we note, was quite laborious.

\subsubsection{Incident Date}
%Incident date is a very important feature of this dataset.
Some of the news dates are written in various formats like ``March 29'', ``June 27, 2018'', ``April 2nd''. But the majority of incident dates found in news articles are mentioned in implicit form such as ``last Friday'', ``yesterday'', ``the day before'', ``before an event'' and so on. In order to determine the actual incident date, we manually adjust these phrases with the news publication dates. All these extracted dates are then manually formatted in "Day-Month-Year" format and included in the dataset.

\subsubsection{Part of the Day of the Incident}
The time when the crime was committed is referred to as the incident time. %It is also an important aspect of this dataset.
Very few news articles state the exact time of the incident. The time of the crime are instead recorded as parts of the day as it is oftentimes difficult to find out the exact time of the crime occurrence. As a result, when categorizing incident time into different parts of the day, we consider the slots/parts of a day shown in Table~\ref{Tab:partofdaytabel} as the crime occurrence time.

\begin{table}[width=.7\linewidth,cols=2,pos=h]
\caption{Deciding the slot/part of the day}
\begin{tabular*}{\tblwidth}{@{} LLLL@{} }
\toprule
Time & Part of the Day\\
\midrule
6 am - 11:59 am     & Morning       \\
12 pm - 3:59 pm     & Noon          \\
4 pm - 5:59 pm      & Afternoon     \\
6 pm - 7:59 pm      & Evening       \\
8 pm - 5:59 am      & Night         \\
\bottomrule
\end{tabular*}
\label{Tab:partofdaytabel}
\end{table}

\subsection{Feature Engineering}
Table~\ref{Tab:derivedfeaturenews} shows the features we derive from the features of Table~\ref{Tab:rawfeaturenews}. Incident place is split into three features, which are: incident division, incident district, and incident place. Additionally, incident dates has been transformed into date-time objects using the Python's Datetime \cite{dateime} module. Then, incident month, incident week, and incident weekday are extracted from these \emph{datetime} objects.

\begin{table}[width=.7\linewidth,cols=3,pos=h]
\caption{Derived Features from three core features of Table~\ref{Tab:rawfeaturenews}.}
\begin{tabular*}{\tblwidth}{@{} LLLL@{} }
\toprule
Feature & Type & Derived From \\
\midrule
Incident Month        & Categorical       & Incident Date     \\
Incident Week         & Categorical       & Incident Date     \\
Incident Weekday      & Categorical       & Incident Date     \\
Weekend               & Categorical       & Incident Date     \\
Part of the Day       & Categorical       & Part of the Day   \\
% Latitude              & Numerical         & Incident Place    \\
% Longitude             & Numerical         & Incident Place    \\
Incident Place        & Categorical       & Incident Place    \\
Incident District     & Categorical       & Incident Place    \\
Incident Division     & Categorical       & Incident Place    \\
\bottomrule
\end{tabular*}
\label{Tab:derivedfeaturenews}
\end{table}

\textbf{Feature Scaling}: %The time and weather features are mostly categorical, with the exception of latitude and longitude. Weather data and demographic data, on the other hand, are numerical in nature. Before applying classification, both feature scaling and feature encoding is required.
Feature scaling is a technique for normalizing the numerical valued features in a fixed range. Without scaling features, the learning algorithm may be biased toward features with greater magnitude values. As a result, feature scaling brings all features into the same range, and thus the learning model makes good use of all features. For numeric features, we apply a well-known scaling technique called min-max normalization. This technique brings every numerical attribute in a defined range. The most common ranges used are [0, 1] and [-1, 1]. The equation for range [-1, 1] is as follows: %Equation for range [-1,1] can be seen here \ref{eq:1}.
\begin{equation*}
\label{eq:1}
\mathrm{x}_{i}^{} = 2 * \frac{\mathrm{x}_{i}^{}-min(x)}{max(x)-min(x)} - 1,
\end{equation*}
where $x$ is the feature at hand and $x_i$ is the value of $x$ at $i$th datapoint.

\textbf{Feature Encoding}: Machine learning algorithms can only operate on numerical values. Hence it is necessary to convert the categorical features into numeric ones. There are various methods for encoding categorical features. We use the label encoding method which turns categorical data into machine-readable numeric form by assigning a unique number (beginning with 0) to each class of a particular feature.

Different types of information are contained in geo-location data, temporal data, weather data, and demographic data. %Geo-location features include location information such as the incident division, district, Upazilla, latitude, and longitude. Temporal features include time and date information such as the incident month, week, weekday, and part of the day of incident. Weather features like max temp, avg temp, min  temp, precip, humidity, visibility, cloud cover, heat index, season, and weather represents weather data. The numerical features in demographic information includes household, male and female population, total population, gender ratio, average size of a home, density per square kilometer, literacy rate, religious institution, playground, park, police station, cyber cafe, school, college, and cinema.
All of these features are combined in the dataset, thereby resulting in a strong set of features for the prediction task. Table~\ref{Tab:allfeature} shows all these features along with some of their properties. Thus the final dataset contains 6674 instances of crimes and 36 features along with a label field indicating the type of the crime.

\begin{table}[width=.85\linewidth,cols=8,pos=h]
\caption{Complete list of features (cf. Tables~\ref{Tab:geo-derivedfeaturenews}, \ref{Tab:weatherfeatures}, \ref{Tab:demofeatures}, and \ref{Tab:derivedfeaturenews}) in the final dataset and some of their properties.}
\begin{tabular*}{\tblwidth}{@{} LLLLLLLL@{} }
\toprule
No. & ID & Feature & Type  & Unique & Range & Median & Information Source\\
\midrule
1 & C1 & Incident Month        & Categorical  & 12  & -              & -     & Crime News   \\
2 & C2 & Incident Week         & Categorical  & 53  & -              & -     & Crime News   \\
3 & C3 & Incident Weekday      & Categorical  & 7   & -              & -     & Crime News   \\
4 & C4 & Weekend               & Categorical  & 2   & -              & -     & Crime News   \\
5 & C5 & Part of the Day       & Categorical  & 5   & -              & -     & Crime News   \\
6 & C6 & Latitude              & Numerical    & -   & 20.87 - 26.48  &  23.86 & Crime News   \\
7 & C7 & Longitude             & Numerical    & -   & 88.14 - 92.44  &  90.27 & Crime News   \\
8 & C8 & Incident Place        & Categorical  & 660 & -              & -     & Crime News   \\
9 & C9 & Incident District     & Categorical  & 64  & -              & -     & Crime News   \\
10 & C10 & Incident Division     & Categorical  & 8   & -              & -     & Crime News   \\
11 & W1 & Max Temp              & Numerical    & -   & 17 - 45        & 33    & Weather API   \\
12 & W2 & Avg Temp              & Numerical    & -   & 16 - 40        & 30    & Weather API   \\
13 & W3 & Min Temp              & Numerical    & -   & 8 - 32         & 25    & Weather API   \\
14 & W4 & Weather Code          & Categorical  & 21  & -              & -     & Weather API   \\
15 & W5 & Precipitation         & Numerical    & -   & 0 - 204.6      & 0.8   & Weather API   \\
16 & W6 & Humidity              & Numerical    & -   & 13 - 97        & 68    & Weather API   \\
17 & W7 & Visibility            & Numerical    & -   & 4 - 10         & 10    & Weather API   \\
18 & W8 & Cloudcover            & Numerical    & -   & 0- 100         & 23    & Weather API   \\
19 & W9 & Heatindex             & Numerical    & -   & 15 - 43        & 33    & Weather API   \\
20 & W10 & Season                & Categorical  & 3   & -              & -     & Weather API   \\
21 & D1 & Household Number      & Numerical    & -   & 4872 - 2030000  & 83300   & BD Cencus   \\
22 & D2 & Male Population       & Numerical    & -   & 11300 - 4930000 & 187000 & BD Cencus   \\
23 & D3 & Female Population     & Numerical    & -   & 11200 - 3970000 & 188000 & BD Cencus   \\
24 & D4 & Total Population      & Numerical    & -   & 22900 - 8910000 & 377000 & BD Cencus   \\
251 & D5 & Gender Ratio          & Numerical    & -   & 84 - 203       & 101   & BD Cencus   \\
26 & D6 & Avg House Size        & Numerical    & -   & 3.58 - 8.42    & 4.44  & BD Cencus   \\
27 & D7 & Population Density    & Numerical    & -   & 23 - 30600     & 1239  & BD Cencus   \\
28 & D8 & Literacy Rate         & Numerical    & -   & 26.7 - 74.6    & 53.9  & BD Cencus   \\
29 & D9 & Religious Institution & Numerical    & -   & 0 - 4289       & 818   & BD Cencus   \\
30 & D10 & Playground            & Numerical    & -   & 0 - 253        & 25    & BD Cencus   \\
31 & D11 & Park                  & Numerical    & -   & 0 - 17         & 1   & BD Cencus   \\
32 & D12 & Police Station        & Numerical    & -   & 0 - 74         & 3   & BD Cencus   \\
33 & D13 & Cyber Cafe            & Numerical    & -   & 0 - 478        & 1   & BD Cencus   \\
34 & D14 & School                & Numerical    & -   & 1 - 242        & 45   & BD Cencus   \\
35 & D15 & College               & Numerical    & -   & 0 - 64         & 8   & BD Cencus   \\
36 & D16 & Cinema                & Numerical    & -   & 0 - 40         & 2     & BD Cencus   \\
\bottomrule
\end{tabular*}
\label{Tab:allfeature}
\end{table}

\section{Data Analysis}
\label{sec:data analysis}
%Prior to using classification algorithms,
Now that the crime dataset is fully developed, we perform some exploratory data analysis to better understand the data. %Exploratory Data Analysis is used to conduct preliminary investigations on data in order to discover patterns, identify anomalies, and generate hypotheses.

\begin{figure}[htbp]
% \begin{figure}
   \centering
%   \begin{tabular}{@{}c@{\hspace{.5cm}}c@{}}
       \includegraphics[page=1,width=.90\textwidth]{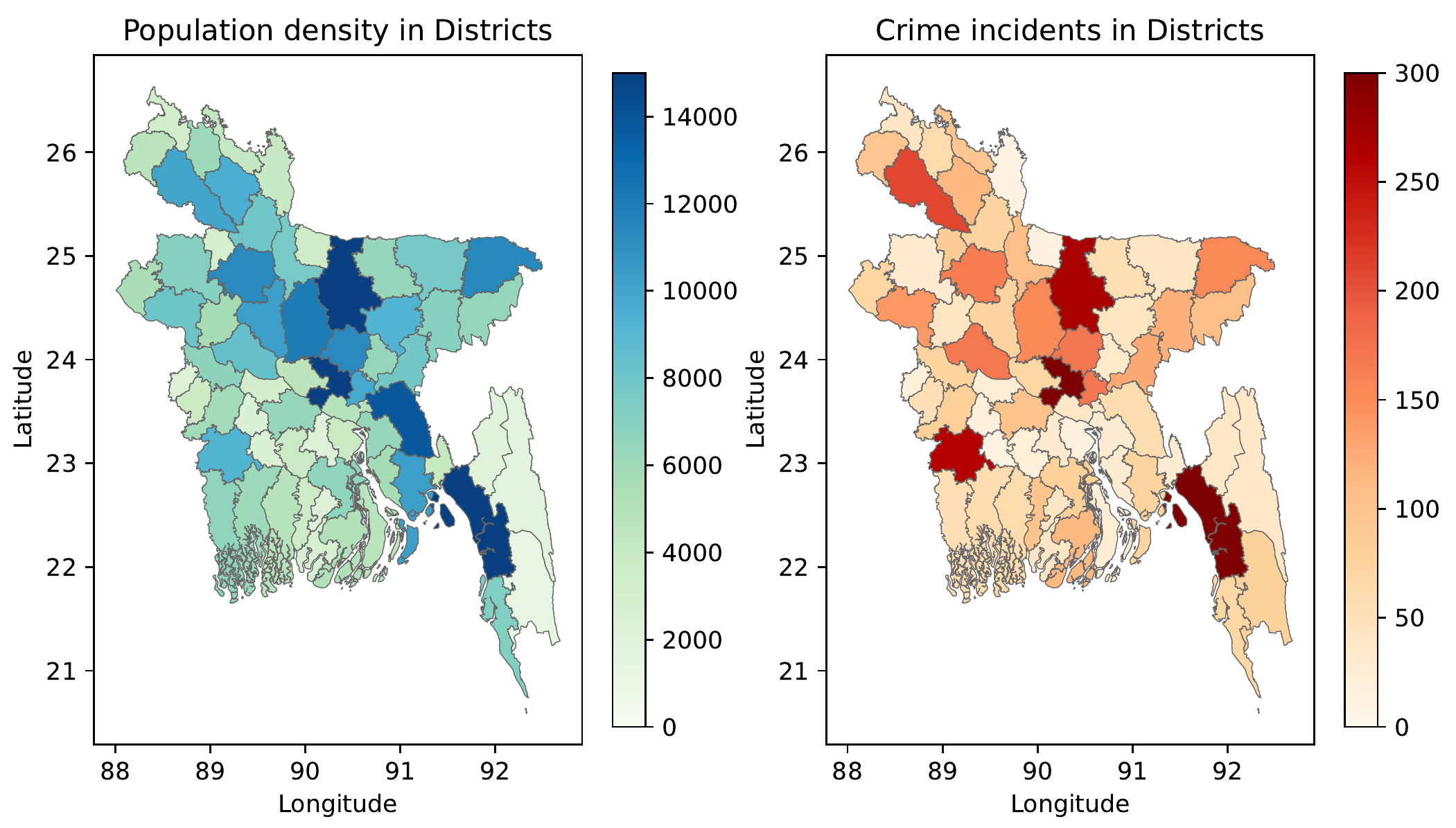}
%   \end{tabular}
 \caption{Population density and crime relation in Districts.}
 \label{fig:densitycrimeheatmap}
\end{figure}

The heatmap shown in Figure~\ref{fig:densitycrimeheatmap} depicts the relationship between population density and crime rate. The general trend is that the crime is more prevalent in the more populous districts.

\begin{figure}[htbp]
% \begin{figure}
   \centering
   \begin{tabular}{@{}c@{\hspace{.5cm}}c@{}}
       \includegraphics[page=1,width=.75\textwidth]{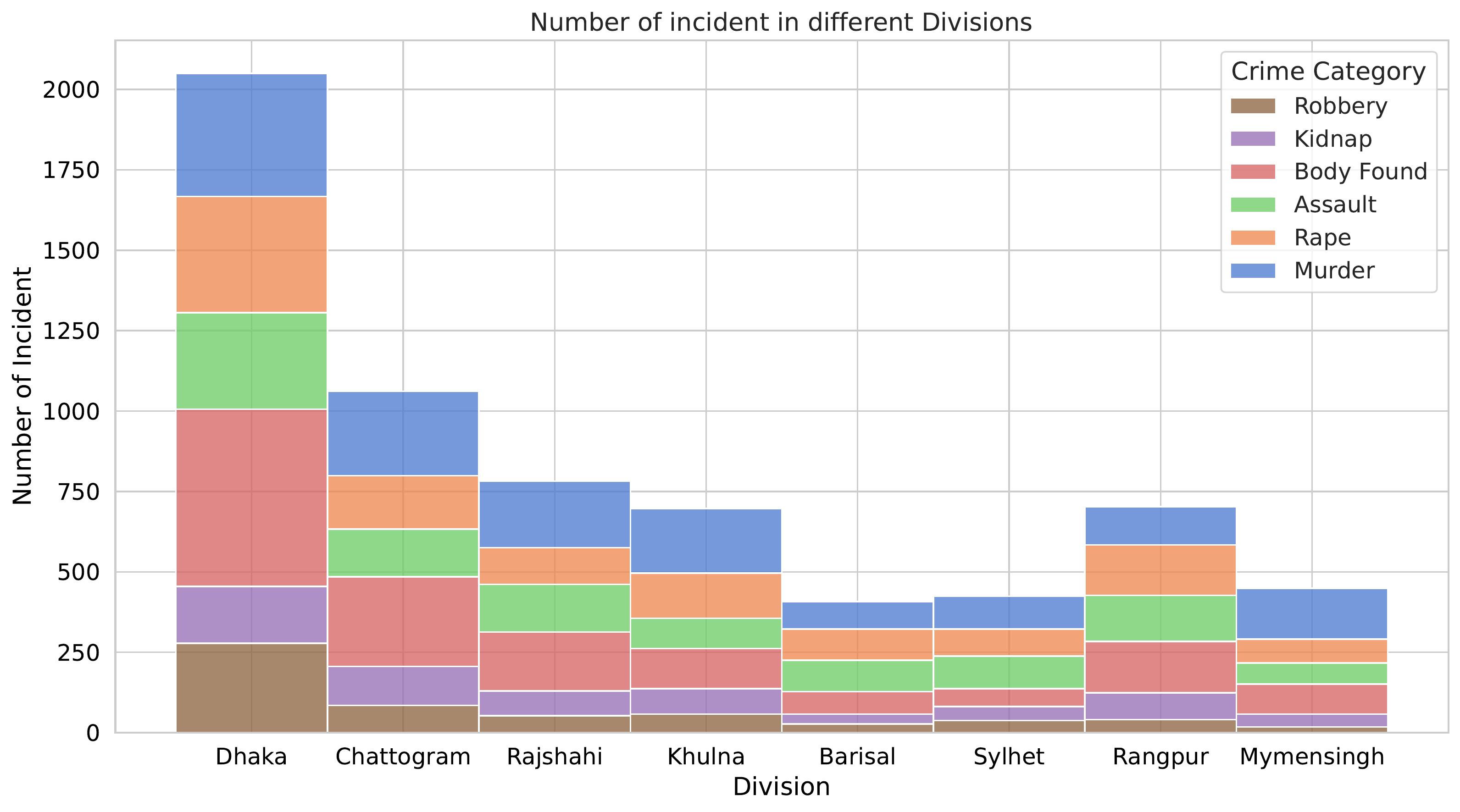}
   \end{tabular}
 \caption{Crime incidents occurring in various Divisions.}
 \label{fig:crimedivision}
\end{figure}

Figure~\ref{fig:crimedivision} illustrates crime incidents in various divisions. It is unsurprising that the most of the recorded crime incidents occurred in the Dhaka division. In addition to having the largest population, Dhaka district, the capital of Bangladesh, is located within this division. We also see that Chattogram has the second highest number of crime incidents, which is perceivable given that it is the second largest city after Dhaka and has the busiest seaport on the Bay of Bengal. It is unusual, however, that despite having a small population, Rangpur has a relatively high rate of crime. This is perhaps due to the fact that the law enforcement infrastructure is poor there, and also that the economic condition of that division is not that good.

\begin{figure}[htbp]
% \begin{figure}
   \centering
   \begin{tabular}{@{}c@{\hspace{.5cm}}c@{}}
       \includegraphics[page=1,width=.65\textwidth]{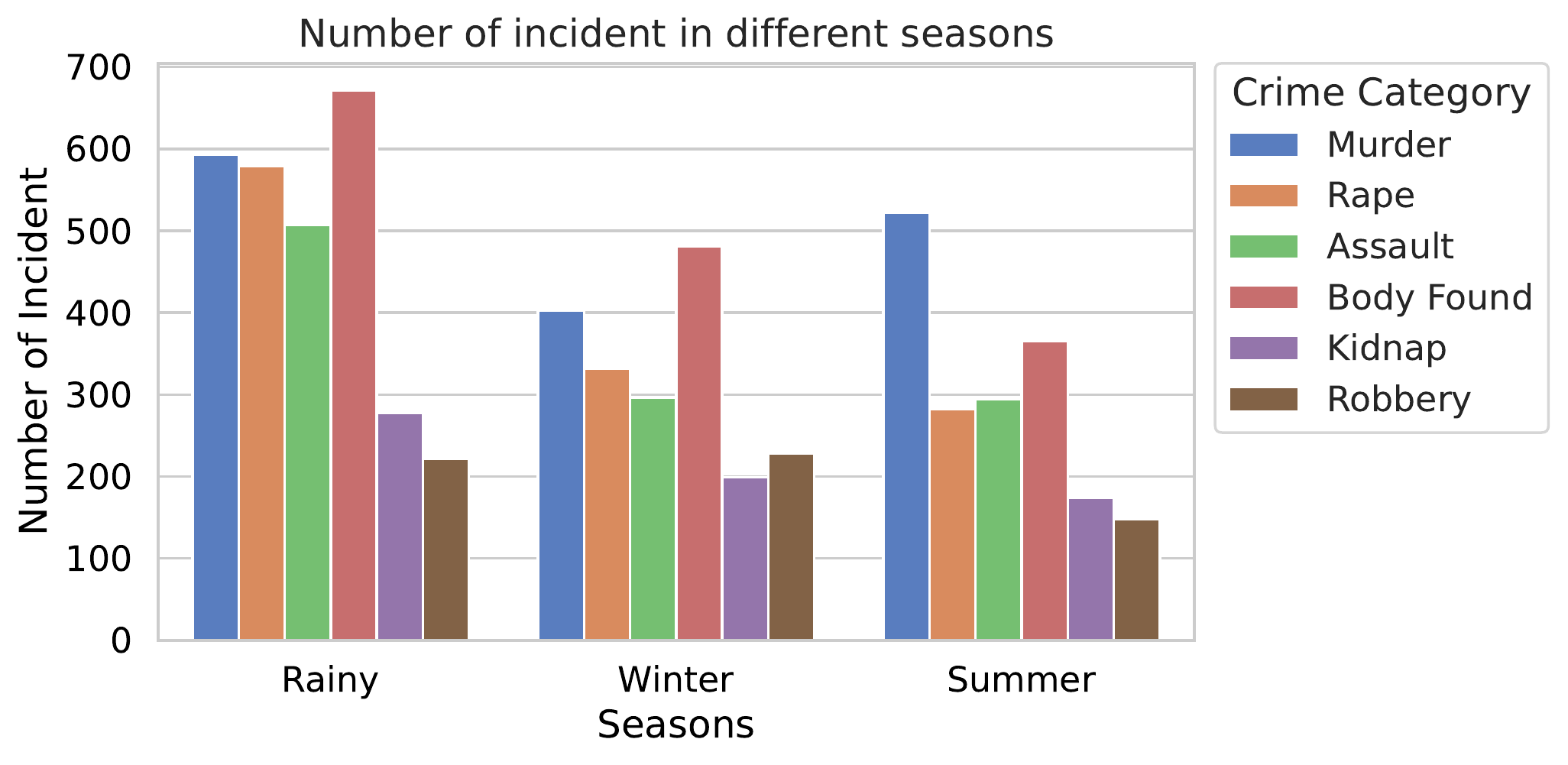}
   \end{tabular}
 \caption{Crime incidents occurring in various seasons.}
 \label{fig:crimeseasons}
\end{figure}

Crime rate in different seasons is visualized in Figure~\ref{fig:crimeseasons}. Broadly, three distinct seasons of Bangladesh are: the hot pre-monsoon summer season which lasts from March through May, the wet monsoon rainy season which lasts from June through October, and the cold dry winter season which lasts from November through February. The Figure shows that the crime occurrence reaches its peak during the rainy season. The murder rate is found to be higher in the summer and rainy season.

\begin{figure}[htbp]
% \begin{figure}
   \centering
   \begin{tabular}{@{}c@{\hspace{.5cm}}c@{}}
       \includegraphics[page=1,width=.95\textwidth]{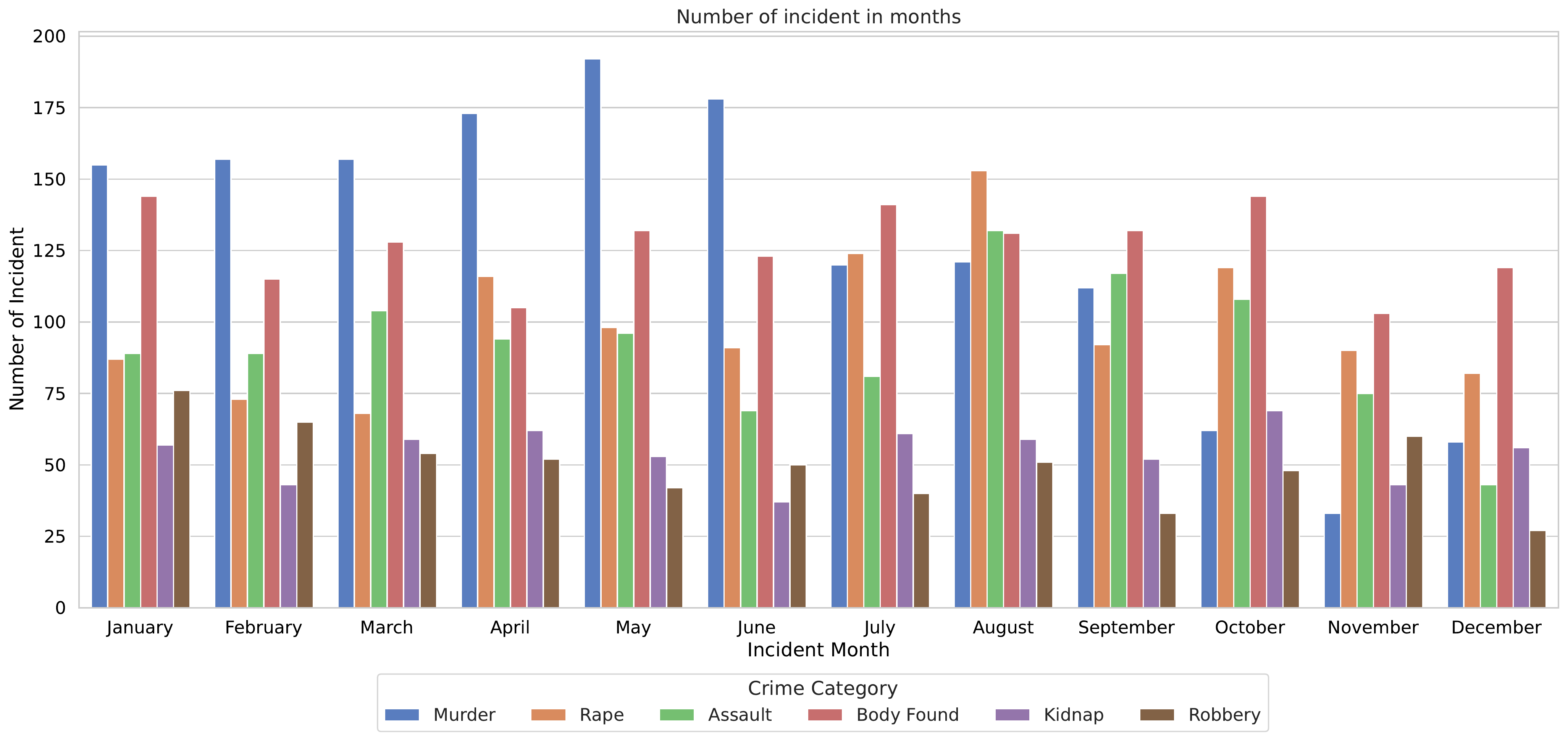}
   \end{tabular}
 \caption{Crime incidents occurring in various months.}
 \label{fig:crimemonths}
\end{figure}

Figure~\ref{fig:crimemonths} shows another significant statistic which is the number of crimes of various categories reported in each month of the year. We see that the number of murders committed each month has a seasonal pattern: the majority of murders were reported in the first six months of the year. On the other hand, from January to May, there was a gradual decrease in the number of robberies. November and December broadly experience low crime rates.

\begin{figure}[htbp]
% \begin{figure}
   \centering
   \begin{tabular}{@{}c@{\hspace{.5cm}}c@{}}
       \includegraphics[page=1,width=0.85\textwidth]{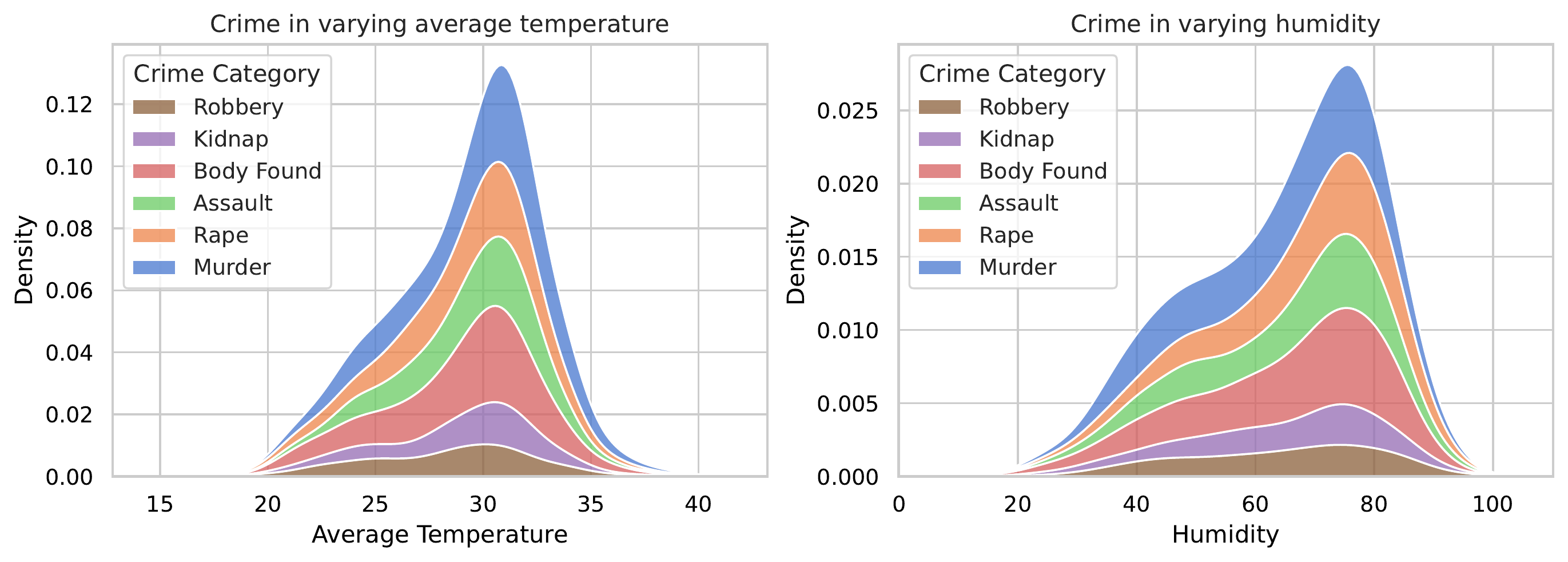}
   \end{tabular}
 \caption{Crime incidents occurring in varying average temperature and humidity.}
 \label{fig:crimeavgtemp}
\end{figure}

As reflected in Figure \ref{fig:crimeavgtemp}, the impact of weather on crime rate is crucial. Crime rates are noticeably higher in hot weather when the average temperature ranges from 28 to 33 degrees Celsius. Also, during the rainy season with a high humidity level, there can be seen an increase in crime incidents.

\begin{figure}[htbp]
% \begin{figure}
   \centering
   \begin{tabular}{@{}c@{\hspace{.5cm}}c@{}}
       \includegraphics[page=1,width=.65\textwidth]{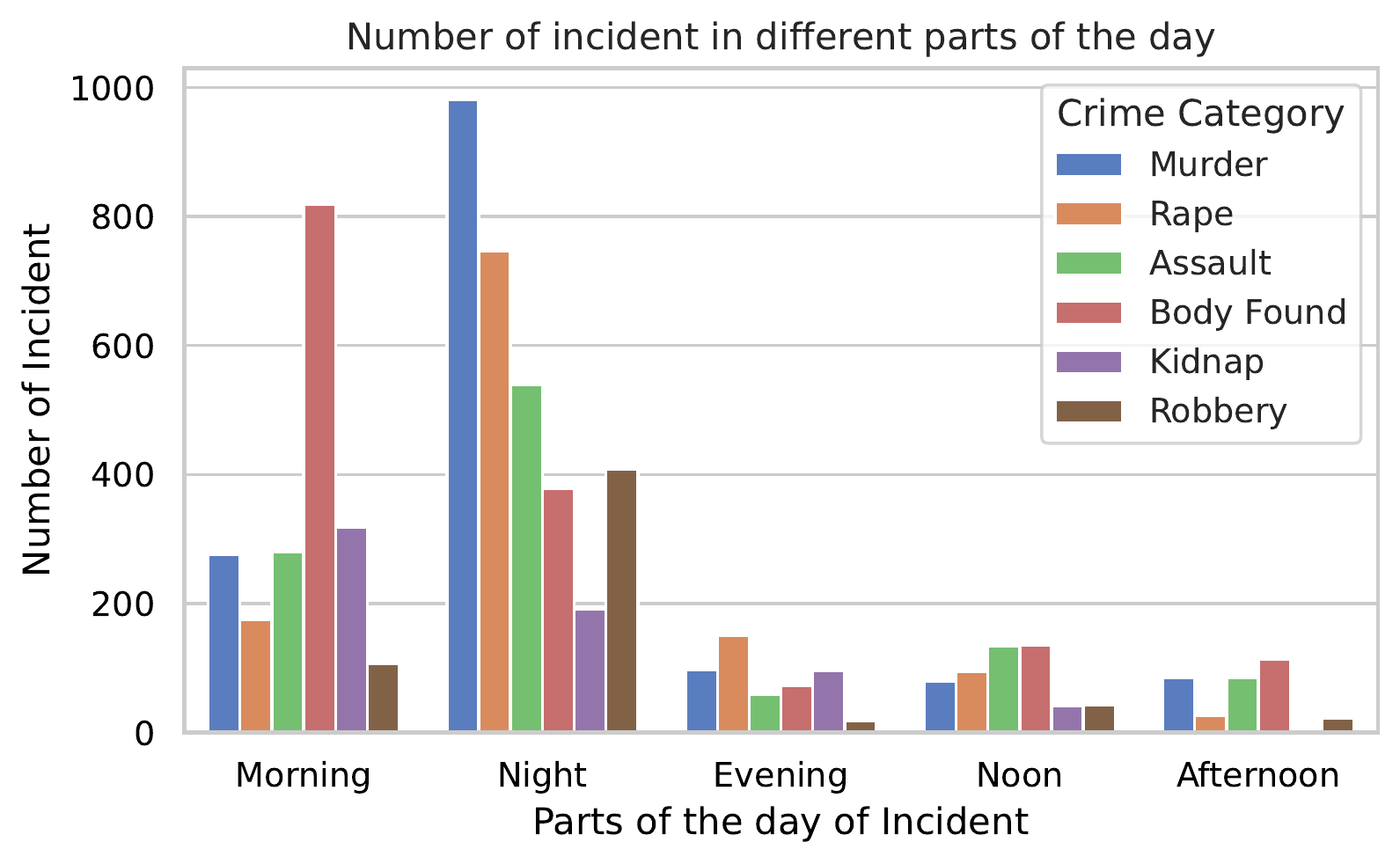}
   \end{tabular}
 \caption{Crime incidents in different part of the day.}
 \label{fig:crimepartoftheday}
\end{figure}

Figure~\ref{fig:crimepartoftheday} demonstrates the trend of the relation between incident time and crime category. Incident time is expressed here as a part of the day.  We see that the majority of murders and rapes occurred at night. The absence of daylight implies that criminals are less exposed which is conducive to criminal activities. The fact that most bodies are discovered in the morning is understandable since most murders that take place at night are found in the morning. %On the other hand, there are generally fewer crimes committed during other times of the day.

\begin{figure}[htbp]
% \begin{figure}
   \centering
   \begin{tabular}{@{}c@{\hspace{.5cm}}c@{}}
       \includegraphics[page=1,width=.70\textwidth]{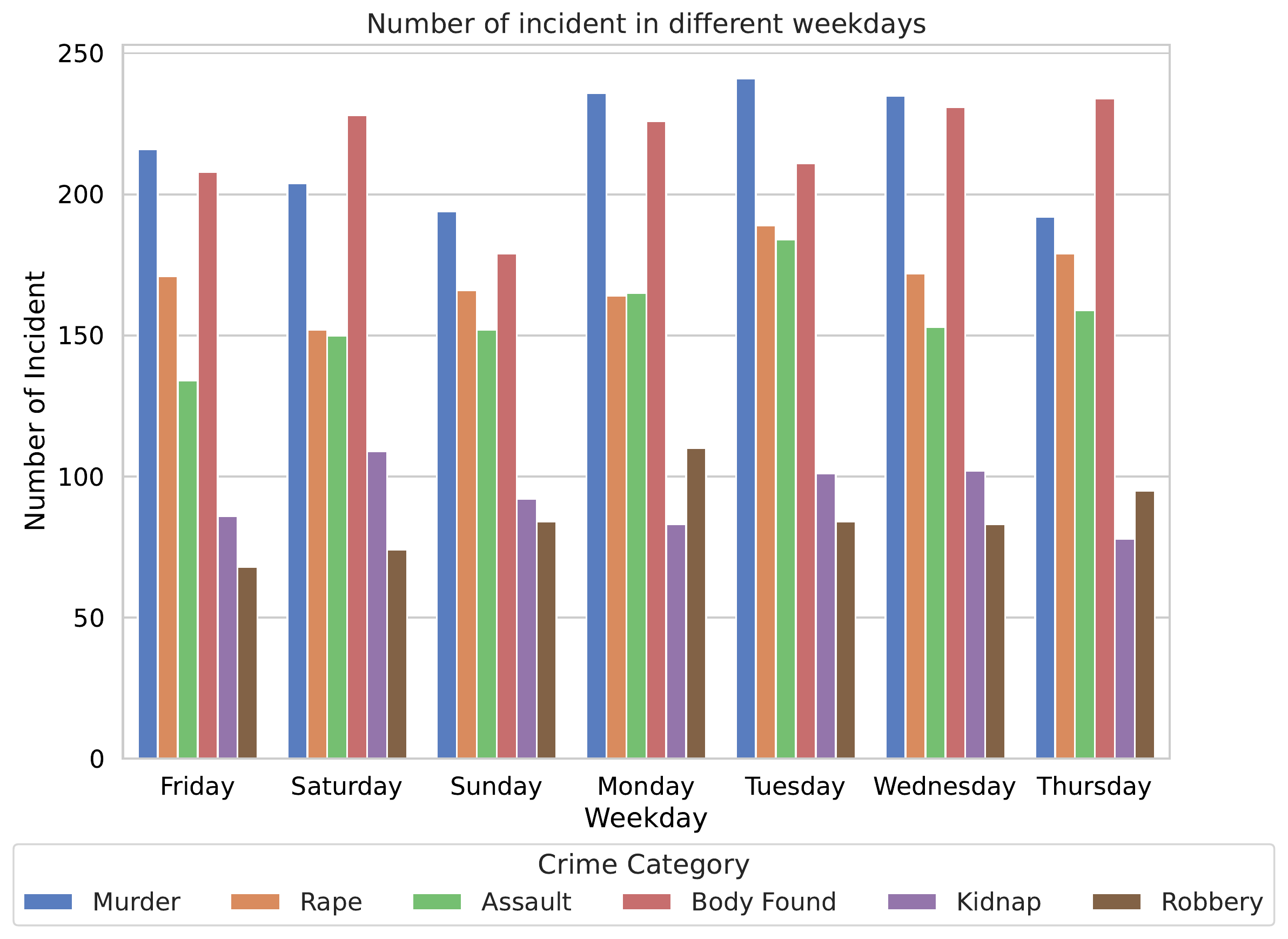}
   \end{tabular}
 \caption{Crime incidents in different weekdays.}
 \label{fig:crimeweekdays}
\end{figure}

Figure~\ref{fig:crimeweekdays} illustrates the frequency of crime incidents during a week. In general, more crimes are recorded during the workdays. From Friday to Sunday, fewer murders occurred as compared to the period of Monday to Wednesday. A trend can be seen that most of the crimes occurred on Tuesdays.

\section{Machine Learning Model Fitting and Result Analysis}
\label{sec:ML models and results}
%The outcomes of each experiment are explored in this section.

We first apply the machine learning models on the original dataset. We then deal with the imbalanced nature of the dataset and conduct experiments on a different setting.

\subsection{Experimental Setup}

We employ five supervised machine learning algorithms that are well-known for their high accuracy, namely Decision Tree, Extra Tree, Random Forest, Adaboost, and Extreme Gradient Boost (XGBoost).

The Decision Tree is a classic classification algorithm where data is split based on a specific feature. Extra Tree Classifier is an ensemble learning technique that aggregates the classification results of multiple de-correlated decision trees. Random Forest is another aggregate classifier that contains a number of decision trees and aggregates their predictions in order to reduce high variance of individual trees. Adaptive Boosting or AdaBoost is a technique that combines multiple weak classifiers to collectively yield a strong classifier. Extreme Gradient Boosting aka XGBoost is a distributed gradient-boosted decision tree.

Decision Tree, Extra Tree, Random Forest, and AdaBoost implementation from ScikitLearn \cite{scikit} library package. For XGBoost, we use another library \cite{xgboost}. For Decision tree, the maximum depth of a tree is set to 20. For Extra Tree, the number of trees is set to 100. For Random Forest, the number of trees is set to 2000, and gini co-efficient is used for splitting. For AdaBoost, the number of trees is set to 50 and learning rate is set to 1.0. For XGBoost, the number of trees, maximum depth, and the learning rate are set to 100, 10, and 0.3 respectively. All other hyperparameters are set to default settings of the libraries. All the experiments are performed on Google Colab \cite{colab} environment.

% XGBoost has 100 trees,  % \textit{n\_estimators} set as 100,
% the learning rate is set to 0.3, and the maximum depth of a tree %\textit{ max\_depth}
% is set to 10. In Decision Tree \textit{max\_depth} was 20, and \textit{criterion} was \textit{gini}. Extra Tree had \textit{n\_estimators} as 100.

Following the standards of machine learning, we split the dataset into two groups: train and test with percentages of 90\% and 10\% respectively. The training set is used for learning (and validation) purpose, and the test set is used for evaluation of the metrics.

%Here, the dataset is split into  traing atwo groups, the train set and the test set, using the percentage split method. The performance of a classifier is determined by comparing the predicted class labels with the actual class labels in the test set.
To assess a classifier's performance, we compute four commonly used metrics, namely, accuracy, precision, recall, and F1 score. ScikitLearn Metrics and scoring library \cite{metrics} is used to calculate these.

\subsection{Experiment on Original Dataset}

Table~\ref{Tab:classifierperf} shows performance of the classifiers in terms of the evaluation metrics. Random Forest's performance tops the list, which is followed by that of Extra Tree and XGBoost. To experimentally stress the need for using a classifier algorithm, we include the performance of random guessing (the last row of the Table) which is found to be quite below all the classifiers. We consider the accuracy satisfactory given that the dataset is developed from a real-world scenario with a lot of missing values and with the presence of noise, and that the task of predicting crime is inherently difficult.

\begin{table}[width=.85\linewidth,cols=5,pos=h]
\caption{Classifiers' performance on original dataset. Results of random guessing is also included.}
\begin{tabular*}{\tblwidth}{@{} LLLLL@{} }
\toprule
Classifier & Accuracy & Precision & Recall & F1\\
\midrule
Random Forest   & 0.44     & 0.44     & 0.44     & 0.43\\
XGBoost         & 0.41     & 0.42     & 0.41     & 0.41\\
Decision Tree   & 0.32     & 0.32     & 0.32     & 0.32\\
Ada Boost       & 0.32     & 0.31     & 0.32     & 0.30\\
Extra Tree      & 0.42     & 0.42     & 0.42     & 0.42\\
Random guess    & 0.14     & 0.14     & 0.14     & 0.14\\
\bottomrule
\end{tabular*}
\label{Tab:classifierperf}
\end{table}

Table~\ref{Tab:classperf} shows the results of Random Forest for every class. We see accuracy of Body-Found is the highest, and accuracy of kidnapping is the lowest.

\begin{table}[width=.85\linewidth,cols=5,pos=h]
\caption{Evaluation of Random Forest for each class with original dataset.}
\begin{tabular*}{\tblwidth}{@{} LLLLL@{} }
\toprule
Class & Accuracy & Precision & Recall & F1\\
\midrule
Assault         & 0.38     & 0.38     & 0.38     & 0.38\\
Body-Found       & 0.52     & 0.42     & 0.52     & 0.47\\
Kidnap          & 0.21     & 0.41     & 0.22     & 0.29\\
Murder          & 0.48     & 0.42     & 0.48     & 0.45\\
Rape            & 0.39     & 0.38     & 0.40     & 0.39\\
Robbery         & 0.25     & 0.47     & 0.25     & 0.33\\
\bottomrule
\end{tabular*}
\label{Tab:classperf}
\end{table}

%\smallskip

%In a later experiment, features such as Weekend, Incident Division, Visibility, Season, Weather, Playground, Partk, Cinema, and Cybercafe were randomly chosen to be removed from the dataset. Hyperparameter were choosen as same as stated before for the classifiers. For the classifiers, all previous hyperparameters were retained. The performance of the classifiers in this experiment is shown in Table~\ref{Tab:classifierperffetdrop}
%\begin{table}[width=.85\linewidth,cols=5,pos=h]
%\caption{Classifiers Performance on imbalanced (original) data after removing random features}
%\begin{tabular*}{\tblwidth}{@{} LLLLL@{} }
%\toprule
%Classifier & Accuracy & Precision & Recall & F1\\
%\midrule
%Random Forest   & 0.41     & 0.42     & 0.41     & 0.41\\
%XGBoost         & 0.40     & 0.40     & 0.40     & 0.40\\
%Decision Tree   & 0.32     & 0.32     & 0.32     & 0.32\\
%Ada Boost       & 0.33     & 0.32     & 0.33     & 0.30\\
%Extra Tree      & 0.41     & 0.41     & 0.41     & 0.40\\
%\bottomrule
%\end{tabular*}
%\label{Tab:classifierperffetdrop}
%\end{table}
%After removing random features, performance drop for Random Forest classifier and Extra Tree classifier is visible here.

%\smallskip

\begin{figure}[htbp]
% \begin{figure}
   \centering
   \begin{tabular}{@{}c@{\hspace{.5cm}}c@{}}
       \includegraphics[page=1,width=0.65\textwidth]{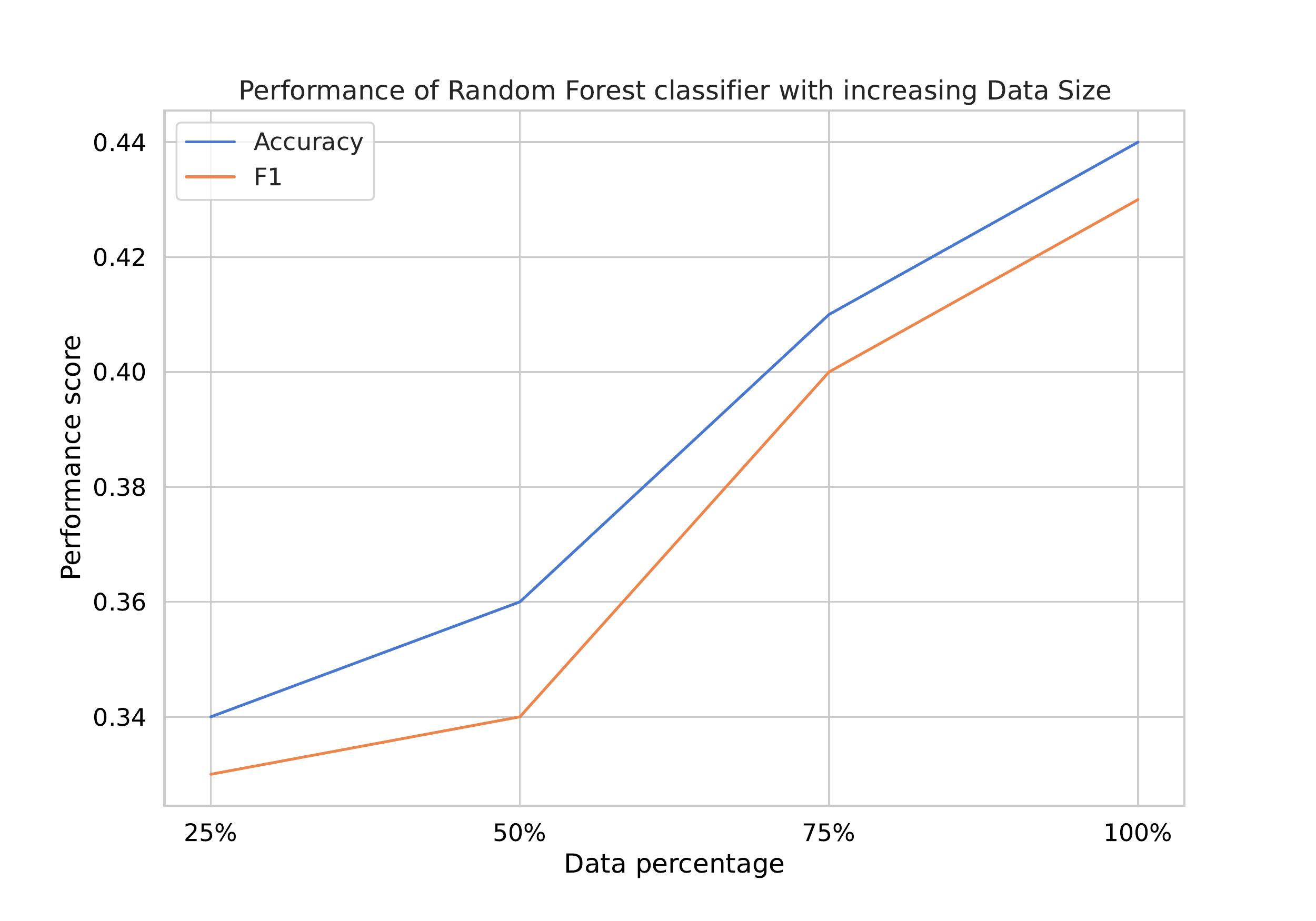}
   \end{tabular}
 \caption{Performance of Random Forest with varying dataset size.}
 \label{fig:peftdataincrease}
\end{figure}

If we increase the dataset size, can we expect a better performance? To investigate this issue, we conduct a pilot experiment with Random Forest where we start with a random 25\% of the dataset, and we then gradually increase the size up to 100\%. % of the dataset starting  Despite the fact that the number of examples in this dataset is quite small, could reducing the number of examples harm classifier performance? Can increasing the size of the dataset gradually, have impact classifier performance? In the following experiment, 25\% of the data were chosen at random. Following that, 50\%, then 75\%, and 100\%.
Figure~\ref{fig:peftdataincrease} shows the plot. From this experiment, we can say that having a larger dataset is likely to yield a better performance.

%\smallskip

%Random Forest classifier's performance on varying size of data is visualized in Figure \ref{fig:peftdataincrease}. This figure can provide answers to those questions. Here as the dataset size grows, so does the performance of classifiers.

From Table~\ref{Tab:classperf} we see that for the two minority classes, namely, kidnapping and robbery, the performance is relatively lower. This raises a question: can we improve the performance by somehow making the dataset more balanced? We try to answer this question next.

\subsection{Experiment on Oversampled Dataset}

\begin{figure}[htbp]
% \begin{figure}
   \centering
   \begin{tabular}{@{}c@{\hspace{.5cm}}c@{}}
       \includegraphics[page=1,width=0.45\textwidth]{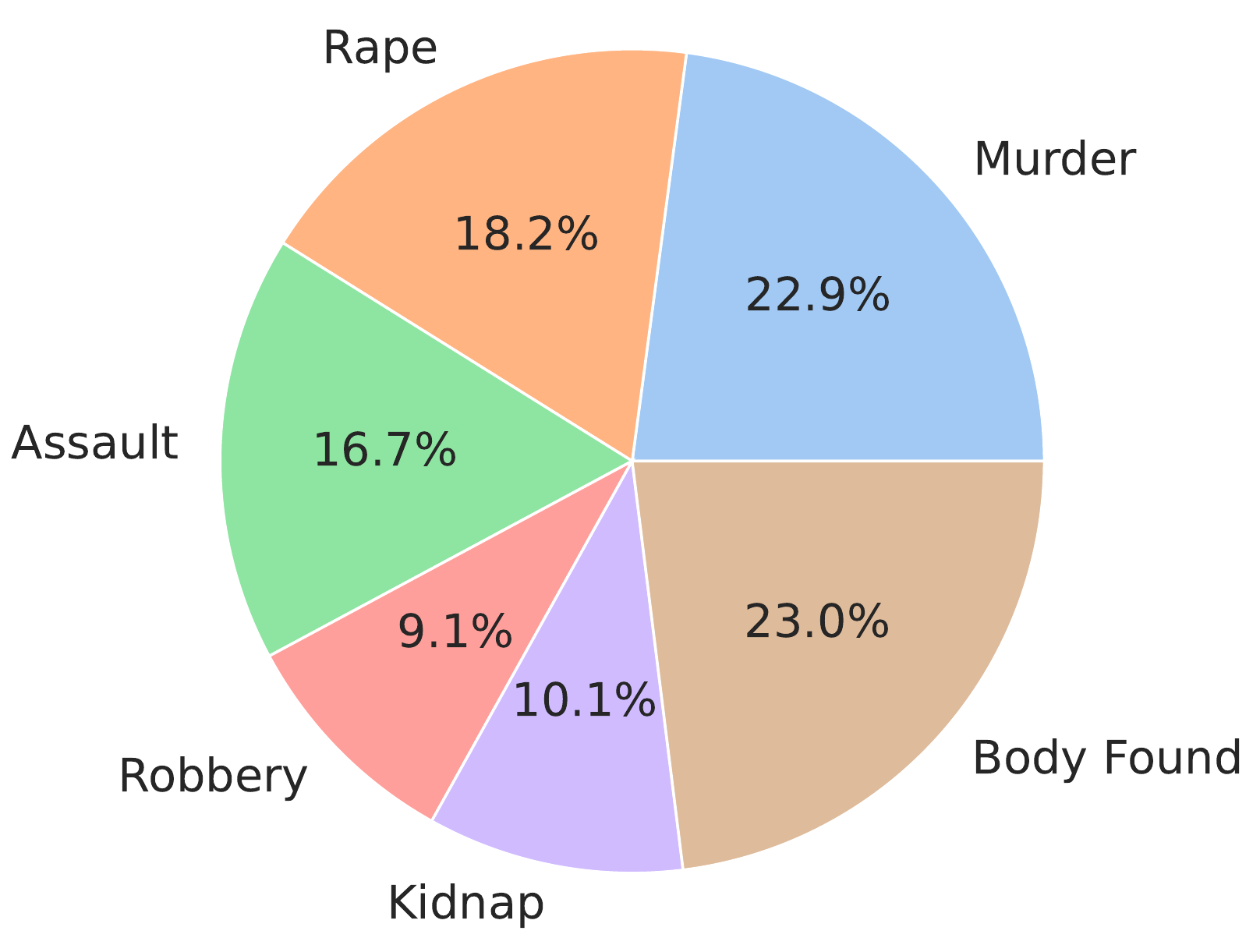}
   \end{tabular}
 \caption{Class distribution of the dataset.}
 \label{fig:crimeclass}
\end{figure}

Figure~\ref{fig:crimeclass} depicts the distribution of classes in the dataset. We see that the distribution is imbalaned: robbery and kidnapping have a much lower percentage of data than other crime categories.

In order to address the problems of an imbalanced dataset, one of the methods that is most frequently chosen by the researchers is resampling the data. There are primarily two types of methods: undersampling and oversampling. Oversampling techniques are typically preferred over undersampling ones because undersampling removes some instances from the data that may contain crucial information. SMOTE \cite{chawla2002smote} is a popular method of oversampling in which artificial samples of the minority classes are systematically generated. We apply this method to our dataset to make it balanced. For this Imbalanced Learn Library \cite{imbalearn} is used. Table~\ref{Tab:datacountBefAftsmote} shows the count of datapoints in each class before and after oversampling with SMOTE.

\begin{table}[width=.85\linewidth,cols=8,pos=h]
\caption{Data points in each class before and after oversampling.}
\begin{tabular*}{\tblwidth}{@{} LLLLLLLL@{} }
\toprule
Dataset & Assault & Body Found & Kidnap & Murder & Rape  & Robbery & Total \\
\midrule
Original        & 1097     & 1517     & 651      & 1518     & 1193     & 598      & 6574 \\
Oversampled-SMOTE    & 1518     & 1518     & 1518     & 1518     & 1518     & 1518     & 9108 \\
\bottomrule
\end{tabular*}
\label{Tab:datacountBefAftsmote}
\end{table}

Table~\ref{Tab:perfsmote} shows classifiers performance after using SMOTE oversampling technique. We see that performance of all algorithms have increased after this change. This time XGBoost and Extra Tree top the list with Random Forest being the second.

\begin{table}[width=.85\linewidth,cols=5,pos=h]
\caption{Classifiers' performance on balanced data after oversampling.}
\begin{tabular*}{\tblwidth}{@{} LLLLL@{} }
\toprule
Classifier & Accuracy & Precision & Recall & F1-Score\\
\midrule
Random Forest   & 0.57     & 0.57     & 0.57     & 0.57\\
XGBoost         & 0.59     & 0.59     & 0.59     & 0.59\\
Decision Tree   & 0.46     & 0.46     & 0.46     & 0.46\\
Ada Boost       & 0.35     & 0.34     & 0.35     & 0.34\\
Extra Tree      & 0.59     & 0.59     & 0.59     & 0.59\\
\bottomrule
\end{tabular*}
\label{Tab:perfsmote}
\end{table}

\subsection{Experiment on Class Reduced Dataset}

%As a result, the experiment was divided into two parts. The imbalanced dataset was used in the first part of the experiments. SOME techniques were also used to deal with imbalanced datasets in this part. However, in the second part of the experiment,
To further investigate the imbalanced data problem, in this experiment we exclude two minority classes -- Robbery and Kidnapping -- altogether to examine if their exclusion improves the overall performance. %Two minority class Robbery and kidnap were excluded from the original dataset in these part of the experiment. The other classes in the dataset were generally balanced. Classifier's performance on these class reduced dataset in given in
Table~\ref{Tab:classifierperfreduced} shows the result. We see that performance in general have increased for all algorithms as compared to the original dataset (cf. Table~\ref{Tab:classifierperf}). Ada Boost notices this increase more than others: the balanced dataset contributes to the highest amount of performance gain for it.

\begin{table}[width=.85\linewidth,cols=5,pos=h]
\caption{Classifiers' performance on class reduced balanced data.}
\begin{tabular*}{\tblwidth}{@{} LLLLL@{} }
\toprule
Classifier & Accuracy & Precision & Recall & F1-Score\\
\midrule
Random Forest   & 0.50     & 0.50     & 0.50     & 0.50\\
XGBoost         & 0.49     & 0.49     & 0.49     & 0.49\\
Decision Tree   & 0.41     & 0.42     & 0.41     & 0.41\\
Ada Boost       & 0.46     & 0.45     & 0.46     & 0.44\\
Extra Tree      & 0.49     & 0.49     & 0.49     & 0.48\\
\bottomrule
\end{tabular*}
\label{Tab:classifierperfreduced}
\end{table}

%After being trained on a more balanced dataset, the classifier's performance significantly improved.

\begin{figure}[htbp]
% \begin{figure}
   \centering
   \begin{tabular}{@{}c@{\hspace{.5cm}}c@{}}
       \includegraphics[page=1,width=0.70\textwidth]{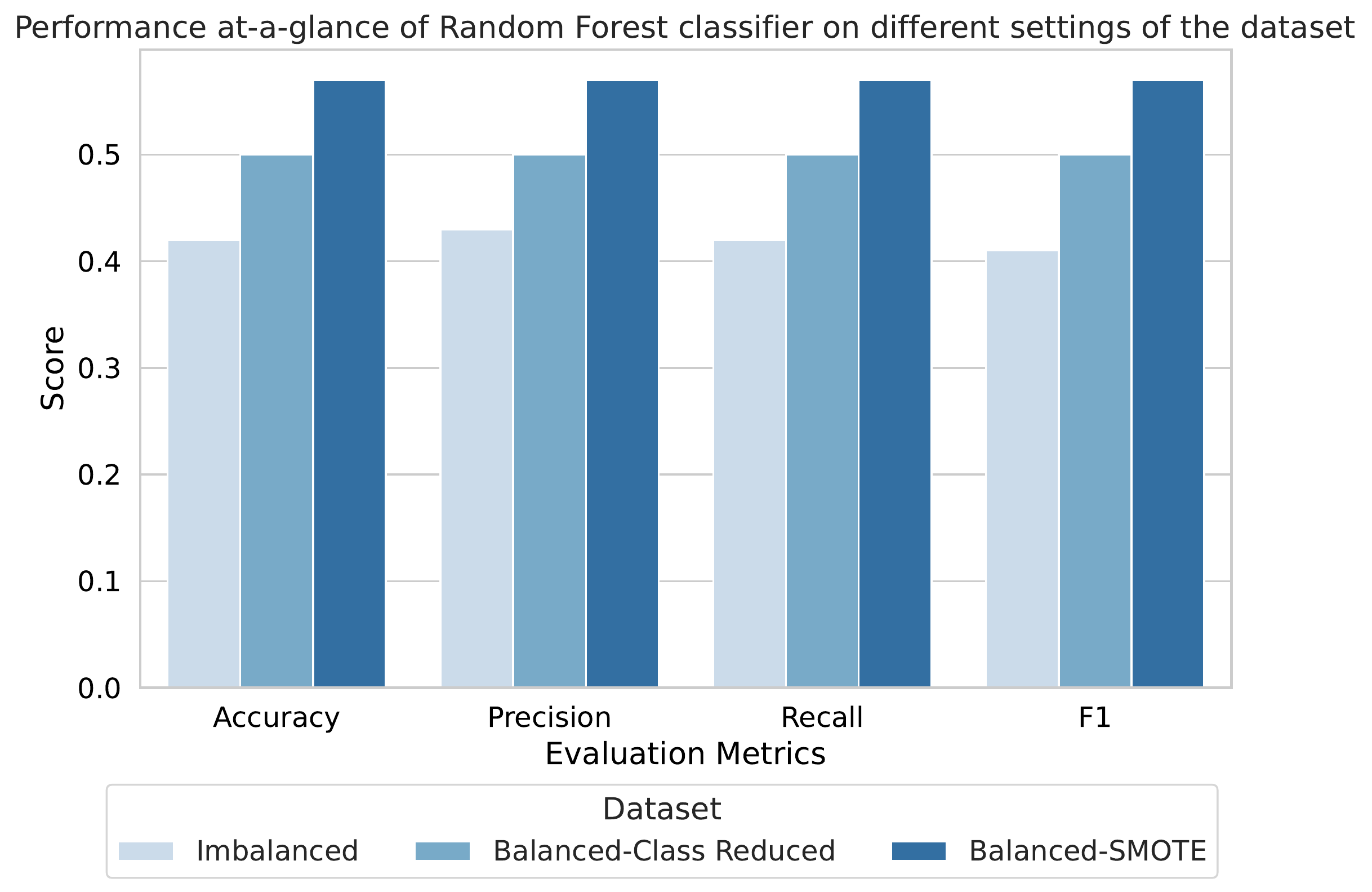}
   \end{tabular}
 \caption{Performance of Random Forest on different settings of the dataset.}
 \label{fig:perdiffverdata}
\end{figure}

When algorithms are trained on more evenly distributed data, performance of the classifier increase in our experiments. Figure~\ref{fig:perdiffverdata} illustrates how Random Forest classifier performs on various settings of the dataset.

% \newpage

In our experiments, we find that Random Forest is the most robust algorithm in the sense that it has been among the top few performs in various settings. This is perhaps because the dataset, being built from a real-world scenario, is inherently noisy; and it is known from the literature \cite{ibrahim2020empirical} that Random Forest is relatively robust to various inconsistencies of the dataset.

\subsection{Experiment on Feature Importance}
The term \emph{feature importance} refers to methods that assign a score to each input feature in a given model indicating the importance of a particular feature on prediction. A higher score indicates that the particular feature has a greater impact on the model's prediction of the target variable.

\begin{figure}[htbp]
% \begin{figure}
   \centering
   \begin{tabular}{@{}c@{\hspace{.5cm}}c@{}}
       \includegraphics[page=1,width=0.80\textwidth]{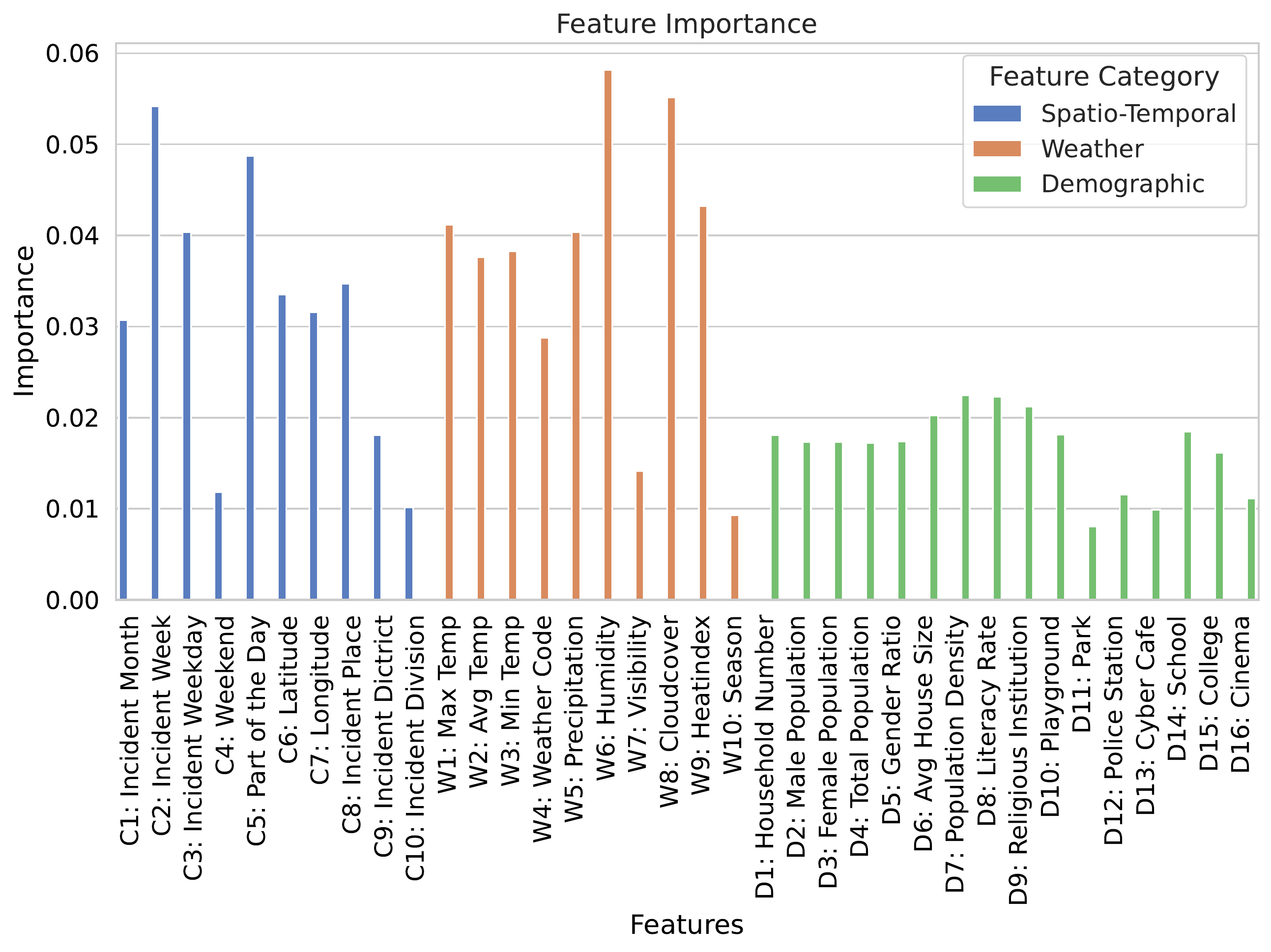}
   \end{tabular}
 \caption{Feature Importance}
 \label{fig:featureimportance}
\end{figure}

Among many available feature importance calculation method, we use the one provided by, or embedded in, the Random Forest classifier. In this method, first the dataset is fit with a Random Forest model. Then, the importance score of a feature is computed as the mean and standard deviation of accumulation of the impurity decrease within each tree. Figure~\ref{fig:featureimportance} presents the feature importance scores of all 36 attributes. We see that the impact of weather data and spatial-temporal crime features on crime prediction is relatively higher. Feature importance of incident week and part of the day are very significant. %, both are categorical attribute that specifies temporal information about the crime committed.
As was noticed in Figure~\ref{fig:crimepartoftheday}, a high number of crimes are committed at night. Weather information such as temperature, humidity, and cloud-cover are also found to be of high feature importance which is in accordance with an existing study by Heilmann and Kahn mentioned earlier \cite{NBERw25961}. Also, in Figure~\ref{fig:crimeavgtemp} we have already demonstrated that the average temperature and humidity have high impact on criminal activities. %However, demographic data have a minor impact on forecasting. Feature importance of season and visibility from weather  information is very low, just like weekend and incident division features from Spatio-Temporal information.

\subsection{Discussion}

%This study elicits a few interesting insights which we describe below.

Leading machine learning scientists and engineers often lament that the use of this astonishing technology is not yet fully leveraged in many important sectors of human society such as agriculture, healthcare, and social well-being.\footnote{\url{https://fortune.com/2019/09/10/a-i-s-next-big-breakthrough-eye-on-a-i/}} The present study is an endeavor to make the benefit of this technology available for social well-being. By collecting, analyzing, and engineering historical crime occurrence data from real-life crime scenarios, we show that it is indeed possible to get crime forecasting -- with a reasonable confidence  -- about future crime occurrence at a particular place and time. %It is common for common people to misinterpret the accuracy as many machine learning models achieve very high accuracy near 100\% for image classification. .... but his is text data so it is known that this tyep of real life data may not have that much accuracy ...

The dataset preparation process for real-time crime scenarios needs to make a number of difficult choices such as the news source selection criteria, the categorization of crime incidents, the type of information to be collected, the process of extracting direct features and deriving indirect ones, the choice of right automated tools etc. While the practices we followed in this study are standard, there are more avenues to explore in almost each of these tasks.

%Choosing the categories of a crime is a contentious point. Not every type of crime can be forecast using historical data, and neither they need to. For example,

Although law enforcement agencies of Bangladesh currently use human common sense to judge the crime-proneness of a particular time and date (for example, crimes are more likely to occur at night as compared to daytime), taking an informed and data-driven decision will definitely boost their success rate in crime prevention. As such, these agencies can reap benefit from studies of this type in a number of ways. First, they get a crime database which, even if no prediction algorithms are used, is inherently valuable for manual data analysis (like the exploratory approach we adopted in Section~\ref{sec:data analysis}). Second, they may deploy a real-time prediction system based on machine learning algorithms in their regional offices and administer their crime patrols accordingly. Third, unlike their current practice of recording crime data in an unplanned and unstructured way (as done in Bangladesh Police website \cite{bdstatisticscrime}) they themselves will be inspired to record the crime incidents following the guidelines of this study.

Although the dataset is prepared from Bangladesh crime incidents, the findings of this study may easily be extended to other countries, especially to the countries with whom Bangladesh has resemblance in terms of demography and weather. Furthermore, using Transfer Learning setting \cite{PanYang2010}, This dataset may also be used for predicting crime of another country having a smaller dataset.

\section{Related Works}
\label{sec:related}

The study of eliciting hidden patterns from crime data using machine learning techniques has been gaining popularity among the researchers since last several years. Below we discuss some relevant works in this field. %Current research indicates that machine learning techniques facilitate the identification of crime patterns.

%NOT RELEVANT: Arulanandam et al. \cite{arulanandam2014extracting} describe how to find out sentences containing crime locations from online news articles. They use Named Entity Recognition (NER) algorithms to identify locations in sentences and Conditional Random Field (CRF) methods to categorize crime locations. This research concentrates on only one type of crime, theft, to demonstrate the viability of such an approach. Four newspaper articles from three different countries are used to evaluate their suggested approach.

Buczak et al. \cite{buczak2010fuzzy}  investigate the use of fuzzy association rule mining to find community crime patterns. They use the Communities and Crime Dataset from the UCI Machine Learning Repository. The data set includes 2215 crime instances and 128 attributes. Almanie et al. \cite{almanie2015crime} concentrate on identifying temporal and spatial criminal hotspots
using a collection of real-world crime statistics of Denver and Los Angeles. Their method is designed to focus on three key aspects of crime data: the type of a crime, when it occurs, and where it occurs. The authors elicit intriguing patterns for crime hotspots using an apriori algorithm. Furthermore, the paper demonstrates how the decision tree and naive bayes classifiers can be used to predict potential crime types. Nguyen et al. \cite{Nguyencrime2017} use Portland Police Bureau's data for crime forecasting. The authors merge the data with demographic information obtained from various public sources. The dataset is then supplemented with additional census data that is accessible for the general people. The entire dataset is used to predict the type of crime in a specific location over time using a variety of machine learning algorithms. The authors assert that the major variables necessary for crime prediction are location and time. They emphasize taking into account the history of crimes as well as other elements including the culprits' demographic, economic, and ethnic characteristics. Bogomolov et al. \cite{Bogomolov2014OnceUA} predict the crime-proneness of a particular area of London city using data collected from mobile phones of users and also using demographic data. %They contend that crime can be predicted by combining basic demographic data and aggregated human behavioral data from the mobile network infrastructure. When predicting whether a particular area of a city will be a crime hotspot or not, their experimental results with real crime data from London achieved an accuracy of nearly 70 percent.

The K-means clustering-based technique is used by Agarwal et al. \cite{agarwalcrime2013} and Buczak et al. \cite{Tayal2015} to analyze yearly crime occurrence patterns.  %These clustering models tends to perform well with categories that have more data.
The dataset used for the former work is crimes recorded by the police in England and Wales from 1990 to 2011. The latter work utilizes the data provided by the National Crime Records Bureau and Committee to Protect Journalists. Varan \cite{nathcrime2007} also examines k-means clustering with a few modifications to assist identification of crime patterns. In order to improve the accuracy of predictions, a semi-supervised learning technique is applied to discover knowledge from crime records. The author develops a weighting scheme for attributes in order to overcome the deficiencies of several out-of-the-box clustering methods. This work applies these techniques to real crime data collected from a local sheriff’s office. Sivaranjani et al. \cite{sivaranjanicrime2016} and Pednekar et al. \cite{Pednekar2018CrimeRP} use some clustering algorithms such as K-means, agglomerative, and DBSCAN to create criminal clusters. The information from the three resulting clusters is then used to predict the class of the crime. The authors obtain data from India's National Crime Records Bureau (NCRB). It contains crime data across six cities, namely Chennai, Coimbatore, Salem, Madurai, Thirunelvelli, and Thiruchirapalli, from 2000 to 2014 with 1760 incidents and 9 attributes. Reddy et al. \cite{toppireddy2018crime} employ Naive Bayes and KNN classifiers to determine the type of crime that is likely to occur using location and day information.  Their data comes from the official website of the United Kingdom Police Department that contains a total of 11 attributes, of which they use crime type, location, date, latitude, and longitude data.

We now discuss some works that apply machine learning and data mining techniques on Bangladesh crime data.

In order to forecast crime trends in Bangladesh, Awal et al. \cite{jakariacrime2016} investigate a linear regression model. The authors use some aggregate data from Bangladesh Police sources. After training the model, crime forecasting for robbery, murder, women and children Repression, kidnapping, and other crimes in the various regions of Bangladesh is attempted. Their experimental findings indicate that the majority of crimes are on the rise as the population increases. Parvez et al. \cite{novel_dhaka_crime} propose a spatio-temporal street crime prediction model that exploits street crime data of Dhaka City. Their dataset is obtained from Dhaka Metropolitan Police (DMP), which consists of the records of crimes from June 2013 to May 2014 but only in aggregate form. %Considering only crime records from the time span of year.
Rahman et al. \cite{rahman_bd_crime}, Islam et al. \cite{islam_bd_crime}, and Biswas et al. \cite{biswas_bd_crime} use Bangladesh Police Statistics \cite{bdstatisticscrime} as the source for their dataset. This dataset contains crime data in a given division over the course of a year, i.e., in aggregate form. It lacks detailed spatio-temporal information. In addition, these authors do not consider demographics in their studies, which is an important feature to predict crime.

From the above discussion we see that although there are some existing works that employ machine learning techniques on Bangladesh crime data, the dataset is neither systematically developed nor comprehensive. Their dataset is in aggregate form, and moreover, lacks sophisticated and effective features such as weather and demographic features. Our work is different from, and more comprehensive, than these works. From this perspective, to the best of our knowledge, our dataset is the first-ever standard crime data for Bangladesh.  We have included not only information from various reliable sources that affect crime occurrence, but also leveraged feature engineering techniques to further improve the quality of the data. Moreover, the size of our dataset can be considered large enough for being eligible to machine learning algorithms.

\section{Conclusion}
\label{sec:conclusion}

The use of machine learning algorithms by criminal analysts can aid in their battle against crime and ensure the safeguard of innocent people. The current study was inspired by the absence of standard, systematic, sizable, and comprehensive crime dataset of Bangladesh. To mitigate this gap, this paper introduces a novel dataset that  includes not only spatio-temporal data, but also weather and demographic data. The dataset is developed using 6574 real-world crime incidents over the duration of 7 years. Exploratory data analysis is performed. State-of-the-art supervised machine learning algorithms are applied to the dataset under different settings and satisfactory results are achieved.

%We discovered an underlying pattern in crime incidents with the help of data analysis on this dataset. The strong relationship between population density and crime incidence, as well as crime frequency in different seasons, is illustrated in section 5. Later, it was come to attention how Dhaka and Chattogram, two of the busiest cities, are plagued by crime incidents.

%Using demographic and spatio-temporal information in the data, this study also sought to predict crime incidents. Supervised algorithm Decision Tree and Ensemble learners such as Random Forest, XGBoost, Ada Boost, and Extra Tree were applied. Random Forest classifiers predicted crime incidents with 44\% accuracy. Since the dataset was imbalanced, the SOMTE Oversampling technique was used later to create a balanced dataset. In this balanced dataset, the XGBoost and Extra Tree classifiers outperformed the other classifiers. Both of these classifiers predicted crime incidents with 59\% accuracy. Another experiment looked at the performance of the Random Forest classifier on a gradually increasing dataset. As does the dataset size increase, so does the performance of the  classifier's.

%The weakness of this study includes the relatively smaller sample size and the imbalanced dataset. The effectiveness of the classifier has been influenced by these factors as well. However, for the time being,
We believe that our research can help the police and other law enforcement organizations predict and prevent crimes that will occur in the future at a specific geographical area. This dataset can serve as a foundation for a database where future crime incidents will be recorded. A national crime reporting system with a huge dataset will be able to better forecast crime incidents. This information may assist law enforcement agencies in better planning and preparing for optimal resource utilization.

%This work can be extended in a number of directions. .. ...

\subsection*{Authors' Contributions}
The contributions of each of the authors are mentioned below: \\
Faisal Tareque Shohan: Planning, investigation, data collection, coding, writing original draft.  \\
Abu Ubaida Akash: Planning, investigation, data collection, writing original draft. \\
Muhammad Ibrahim: Planning, investigation, writing original draft, reviewing original draft, supervision. \\
Md. Shafiul Alam: Planning, investigation, reviewing original draft, supervision.

% \bibliographystyle{cas-model2-names}
%\bibliographystyle{plain}
\bibliographystyle{unsrtnat}
\bibliography{cas-refs}

\end{document}